\def\paperlanguage{}
\pdfoutput=1 

\documentclass[letterpaper, 10 pt, conference]{ieeeconf}  

\usepackage{bm}
\usepackage{cite}
\usepackage{flushend}
\include{preamble}

\IEEEoverridecommandlockouts                              

\title{\LARGE \textbf
  {
    \switchlanguage%
    {%
      Estimation and Control of Motor Core Temperature\\with Online Learning of Thermal Model Parameters:\\Application to Musculoskeletal Humanoids
    }%
    {%
      温度モデルパラメータのオンライン学習を用いたモータコア温度推定と制御: 筋骨格ヒューマノイドへの適用
    }%
  }
}

\author{Kento Kawaharazuka$^{1}$, Naoki Hiraoka$^{1}$, Kei Tsuzuki$^{1}$, Moritaka Onitsuka$^{1}$, Yuki Asano$^{1}$, Kei Okada$^{1}$, \\Koji Kawasaki$^{2}$,  and Masayuki Inaba$^{1}$
  \thanks{$^{1}$ The authors are with the Department of Mechano-Informatics, Graduate School of Information Science and Technology, The University of Tokyo, 7-3-1 Hongo, Bunkyo-ku, Tokyo, 113-8656, Japan.
    {\texttt\small [kawaharazuka, hiraoka, tsuzuki, onitsuka, asano, k-okada, inaba]@jsk.t.u-tokyo.ac.jp}
  }
  \thanks{$^{2}$ The author is associated with TOYOTA MOTOR CORPORATION.
    {\texttt\small koji\_kawasaki@mail.toyota.co.jp}
  }
}
\begin{document}

\maketitle
\thispagestyle{empty}
\pagestyle{empty}

\begin{abstract}
  \switchlanguage%
  {%
    The estimation and management of motor temperature are important for the continuous movements of robots.
    In this study, we propose an online learning method of thermal model parameters of motors for an accurate estimation of motor core temperature.
    Also, we propose a management method of motor core temperature using the updated model and anomaly detection method of motors.
    Finally, we apply this method to the muscles of the musculoskeletal humanoid and verify the ability of continuous movements.
  }%
  {%
    モータ温度の推定・管理はロボットの継続的な動作において極めて重要である.
    本研究では, 正確なモータコア温度の推定のための温度モデルパラメータのオンライン学習について提案する.
    また, そのモデルを用いた筋張力制御による筋温度管理手法, 筋の異常検知について提案する.
    最後に, 本研究を筋骨格ヒューマノイドの全身の筋に適用することで, 継続的な動作が行えることを確認する.
  }%
\end{abstract}

\section{INTRODUCTION}\label{sec:introduction}
\switchlanguage%
{%
  Managing motor temperature is important for the continuous movements of robots.
  Especially, the equipment of motors with sufficient specifications is difficult for the life-sized humanoid robot in order to keep its human-like proportion and weight.
  Thus, joint torque must be limited to keep the motor temperature within the rated value.

  Several management methods of motor temperature have been developed from a variety of perspectives so far.
  As a method to manage motor temperature by hardware, a water cooling method of motors has been proposed \cite{urata2010design} and the developed robot can jump high with a momentary large current.
  Also, motor cooling methods using a heat sink, the air, and phase change materials have been proposed \cite{sevinchan2018thermal}.
  However, these additional hardware is limited for the humanoid robots in order to keep their human-like specifications.

  Common methods to constrain joint torque are minimizing joint torque by optimization \cite{erez2013mpc} and using inequality constraints to set the maximum value of joint torque when optimizing \cite{diehl2006shooting}.
  However, because these methods do not directly handle motor temperature, they cannot guarantee that motor temperature is lower than the rated value.
  Although there is a method estimating motor housing temperature and moving the center of gravity for thermal relaxation \cite{noda2014thermal} as an example to directly consider the motor temperature in the cost function at optimization, the cost function also cannot guarantee the rated value.

  Compared with these methods, Urata, et al. have estimated motor core temperature from motor housing temperature and electric current, and developed a method to calculate the maximum current that can flow during an extremely short amount of time, and constrain a current by the maximum value \cite{urata2008thermal}.
  Also, Kumagai, et al. have proposed a method to estimate motor housing temperature and constrain joint torque by the maximum value which can be applied for 120 seconds \cite{kumagai2014highload}.

  However, there are several problems in the proposed software methods to control the motor temperature.
  First, although there are some parameters for the thermal estimation of motors, among these parameters, the assumption that the ambient temperature is constant is often wrong.
  Regarding motors in the body parts that accumulate heat easily, the ambient temperature changes dynamically.
  Also, thermal and terminal resistances of motors gradually change due to the poor use of the robot or deterioration over time.
  Although identification of thermal model parameters is conducted in \cite{kumagai2014highload}, the identification is conducted offline and only motor housing temperature is considered by simplifying the thermal model.
  Second, there is a problem in the control of motor temperature.
  The proposed methods so far have restricted motor output by the maximum current or joint torque that guarantees the rated value when applying the constant current or joint torque during a certain period.
  Because \cite{urata2008thermal} considers an extremely short amount of time, although it can be applied to instantaneous motions like a jump, it is not suitable for ordinary motions.
  Because \cite{kumagai2014highload} does not change the maximum output dynamically, a value lower than the actual possible value is used as the maximum value.
}%
{%
  モータ温度の管理は, ロボットの継続的な動作において極めて重要である.
  特に, 等身大ヒューマノイド\cite{hirai1998asimo, hirukawa2004hrp}は, 人間と同じような身長・体重を保つため, 全ての動作に十分なモータを搭載することは難しい.
  そのため, 関節トルクを制限し, 温度を定格内に収める必要がある.

  これまで, 様々な方面からモータの温度管理を行う手法が考案されてきた.
  ハードウェアの工夫によりモータ温度を管理する手法として, モータを水冷することで温度の上昇を抑え, 一瞬の大電流によりジャンプが可能なヒューマノイドが開発されている\cite{urata2010design}.
  この他にも, ヒートシンクや空冷, 相転移物質を用いたモータの温度管理方法も提案されている\cite{sevinchan2018thermal}.
  しかし, 等身大ヒューマノイドが人間と同じような身長・体重を保つためには, ハードウェアの工夫には限界がある.

  関節トルクを抑える制御として一般的な方法は, 最適化の際に関節トルクを最小化する\cite{erez2013mpc}, または最適化の際の不等号制約として関節トルクの最大値を設定する\cite{diehl2006shooting}, といったものである.
  しかし, これらは直接的に温度を扱っているわけではないため, 温度が定格より小さくなっていることを保証するものではない.
  また最適化の際のコスト関数として温度を考慮した例として, モータハウジング温度の推定と温度緩和のための重心移動制御\cite{noda2014thermal}が存在するが, コスト関数は温度が定格よりも小さくなっていることを厳密に保証するものではない.

  これに対して, 浦田らはモータハウジング部の温度と電流からモータ内部のコア温度を推定し, コアの温度が定格よりも大きくならないような, ごく微小な時間流し続けられる最大の電流を計算して電流制限を行う手法を開発している\cite{urata2008thermal}.
  また, 熊谷らはモータハウジングの温度推定により120秒間発揮し続けられる最大の関節トルクによって制限を施した制御\cite{kumagai2014highload}を提案した.

  しかし, これまでに提案されたソフトウェアによる温度制御にはいくつかの問題点がある.
  まず, 温度推定のパラメータはいくつか存在するが, このうち, 外部温度が一定という仮定は往々にして間違っている.
  特に, 屋外で夏と冬に実験するのでは, 温度上昇の仕方は全く異なる.
  また, 熱が溜まりやすいような部位のモータでは, 外部温度はDynamicに変化する.
  その上, モータに関する熱抵抗や端子間抵抗も, 劣悪なモータの使い方や経年劣化等によって徐々に変化していくものであり, これも一定とは限らない.
  \cite{kumagai2014highload}では温度推定パラメータの同定を行ってはいるが, オフラインの実行であり, また, 温度モデルを簡易化しているためモータハウジング温度しかわからない.
  次に, 温度制御の方法である.
  これまで述べた手法では, ある一定期間同じ電流/関節トルクを発揮したときに, 定格温度を上まらない最大の電流/関節トルクを使ってそれらを制限していた.
  \cite{urata2008thermal}はごく微小な時間を考えているため, ジャンプのような瞬発的な動作には適用可能だが, 通常の動作では突然最大電流値が下がりロボットが停止してしまう.
  \cite{kumagai2014highload}では電流/関節トルクを動的に変えていないため, 本当に出力可能な最大値よりも小さな値を最大値として使用することになってしまう.
}%

\begin{figure}[t]
  \centering
  \includegraphics[width=1.0\columnwidth]{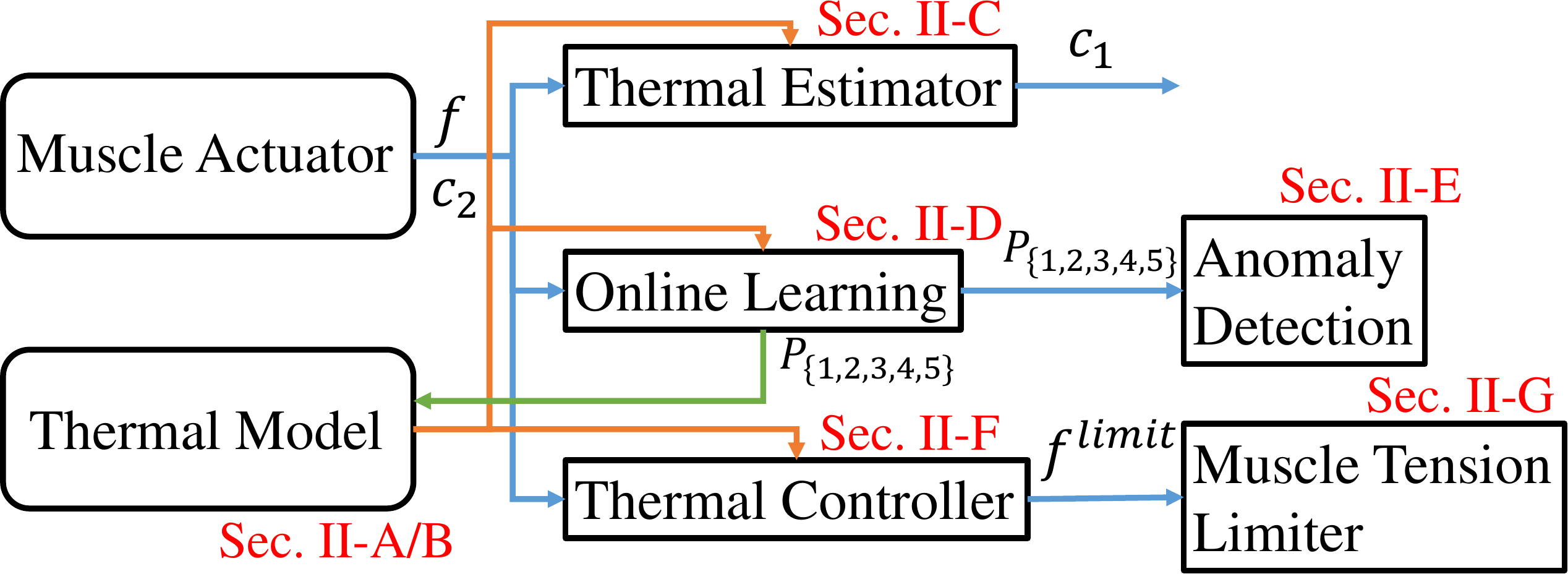}
  \vspace{-3.0ex}
  \caption{System overview.}
  \label{figure:system-overview}
  \vspace{-3.0ex}
\end{figure}

\switchlanguage%
{%
  Therefore, we propose an online learning method of thermal model parameters for the estimation of accurate motor core temperature, and the accurate control of the temperature using the updated model and dynamic optimization.
  Also, we conduct anomaly detection of motors using the change in thermal model parameters.
  In this study, we verify the effectiveness by applying this method to the musculoskeletal humanoid.
  Because the musculoskeletal humanoid \cite{gravato2010ecce1, kawaharazuka2019musashi} mimics the human body in detail and has many muscles, the restrictions of weight and proportion are strict, and air or water cooling of motors is difficult to apply (on the other hand, there is a study on sweating robots \cite{kozuki2016sweat}).
  Since the improvement of hardware is difficult, thermal restriction by software is suitable.
  The actuator of the musculoskeletal humanoid used in this study is not a pneumatic actuator but a motor which winds a muscle wire by a pulley.
  When applying this study to ordinary axis-driven humanoids, muscle tension should be converted to joint torque.

  The contributions of this study are shown below.
  \begin{itemize}
    \item Online learning of thermal model parameters for an accurate estimation of motor core temperature
    \item Dynamic calculation of maximum output of motors using the updated thermal model and optimization
    \item Anomaly detection of motors using the change in thermal model parameters
    \item Verification of continuous motions of the musculoskeletal humanoid by applying this method to a muscle length-based control
  \end{itemize}
}%
{%
  そこで本研究では, \figref{figure:system-overview}に示すような, モータコア温度推定のパラメータのオンライン学習・またそのモデルを用いた最適化による動的な最大出力制限の制御を行う.
  また, パラメータ変化を用いたモータの異常検知等も行う.
  本研究では, 本制御を筋骨格ヒューマノイドの全身の筋に適用することで, その有効性を確かめる.
  筋骨格ヒューマノイド\cite{gravato2010ecce1, nakanishi2013design, kawaharazuka2019musashi}は, 人体を詳細に模倣しているがゆえ, 多数の筋を有し, 重量やプロポーションの制限が厳しく, 空冷や水冷が現状では難しい(汗をかく手法も開発されている\cite{kozuki2016sweat}).
  そのため, ハードウェアの改造が難しく, ソフトウェアによる温度制限が適している.
  本研究で扱う筋骨格ヒューマノイドは, 筋が空気圧等ではなく, モータとプーリによって紐を巻き取る方式のものについて考える.
  本手法を軸駆動型ヒューマノイドに適用する際は, 筋張力を関節トルクに変換して同様に適用することが可能である.

  本研究の貢献は以下である.
  \begin{itemize}
    \item モータコア温度推定のための温度モデルパラメータのオンライン学習
    \item 温度モデルを用いた動的な最大出力制限値の計算
    \item 温度モデルパラメータ変化を使ったモータ異常検知
    \item 筋骨格ヒューマノイドの筋長制御への適用による継続的動作の確認
  \end{itemize}
}%

\begin{table}[t]
  \centering
  \caption{Notations in this paper}
  \label{table:notations}
  \begin{tabular}{c|c}
    Notation & Definition \\
    \hline \hline
    $c_{1}$ & motor core temperature [$^\circ$C]\\
    $c^{max}_{1}$ & maximum motor core temperature to be limited [$^\circ$C]\\
    $c_{2}$ & motor housing temperature [$^\circ$C]\\
    $c_{a}$ & ambient temperature [$^\circ$C]\\
    $f$ & muscle tension [N]\\
    $f^{limit}$ & maximum muscle tension to be limited [N]\\
    $l^{ref}$ & target muscle length [mm]\\
    $\Delta{l}$ & muscle elongation value [mm]\\
    $\bm{\bullet}_{k}$ & the value at the time step $k$ \\
    $\bm{\bullet}_{[k^{from}, k^{to}]}$ & the value sequence at the time steps from $k^{from}$ to $k^{to}$ \\
    $P_{\{1, 2, 3, 4, 5\}}$ & thermal parameters to be updated \\
    $P_{\{1, 2, 3, 4, 5\}, sim}$ & thermal parameters of simulated muscle actuator \\
  \end{tabular}
\end{table}

\section{Proposed Method} \label{sec:proposed}
\switchlanguage%
{%
  First, we will explain a basic two-resistor thermal model to estimate motor core temperature.
  After that, we will construct a model with thermal parameters as variables, and explain thermal estimation of motors, online learning of the thermal parameters, anomaly detection of motors, calculation of maximum output, and control method to restrict muscle tension.
  The entire system and corresponding sections are shown in \figref{figure:system-overview}, and notations are shown in \tabref{table:notations}.
}%
{%
  まず, モータのコア温度を推定する際の基本的な2抵抗モデルについて述べる.
  その後, それらのパラメータを変数とするモデルを構築し, 温度推定・パラメータのオンライン学習・異常検知・温度制限制御・筋張力制限制御について述べていく.
}%

\begin{figure}[t]
  \centering
  \includegraphics[width=0.5\columnwidth]{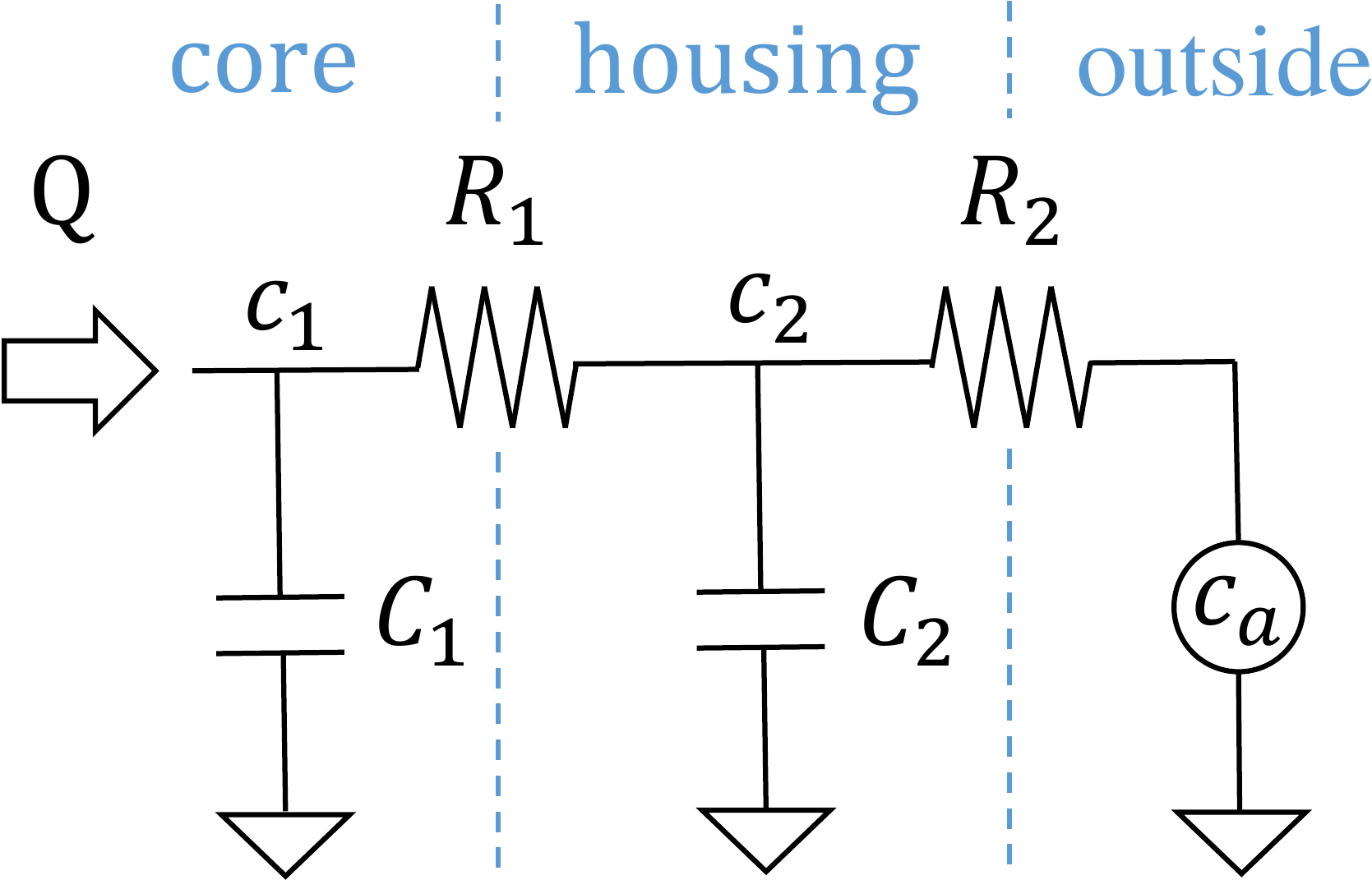}
  \caption{Overview of a basic two-resistor thermal model.}
  \label{figure:basic-model}
  \vspace{-3.0ex}
\end{figure}

\subsection{Basic Thermal Model}
\begin{figure}[t]
  \centering
  \includegraphics[width=1.0\columnwidth]{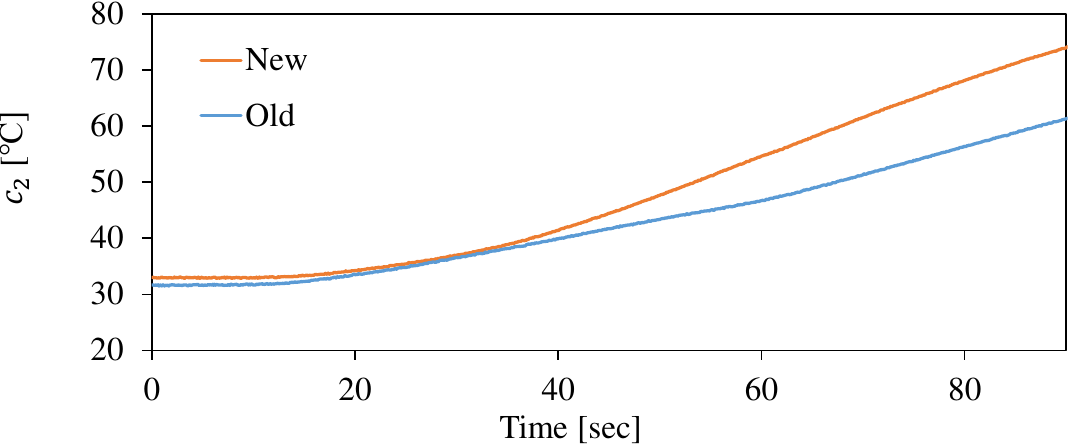}
  \caption{Difference of transition of $c_{2}$ when applying $f=200$ [N] between new and old motors}
  \label{figure:thermal-difference}
  \vspace{-3.0ex}
\end{figure}

\switchlanguage%
{%
  In this study, we use a two-resistor thermal model as shown in \figref{figure:basic-model}, which is the same model as stated in \cite{urata2008thermal}.
  This model is applicable to various classical motors such as brushless and brushed DC motors.
  We assume heat capacity $C_{1}, C_{2}$ for motor core and housing, thermal resistance $R_{1}$ between motor core and housing, and thermal resistance $R_{2}$ between motor housing and ambient temperature.
  There is a relationship among $c_{1}$, $c_{2}$, and $c_{a}$, as shown below,
  \begin{align}
    C_{1}\frac{dc_{1}}{dt} &= Q - \frac{c_{1}-c_{2}}{R_{1}} \label{eq:basic-1}\\
    C_{2}\frac{dc_{2}}{dt} &= \frac{c_{1}-c_{2}}{R_{1}} - \frac{c_{2}-c_{a}}{R_{2}} \label{eq:basic-2}\\
    Q &= R_{e}i^{2}
  \end{align}
  where $R_{e}$ is a wire-wound resistance and $i$ is an electric current that flows through the motor core.
  Also, in order to apply this model to the musculoskeletal humanoid, we convert $i$ to $f$ by the equation below,
  \begin{align}
    D_{pulley}f =E_{gear}D_{gear}E_{motor}K_{t}i
  \end{align}
  where $K_{t}$ is a torque constant, $E_{\{motor, gear\}}$ is a transmission efficiency of the motor or gear, $D_{gear}$ is a gear ratio, and $D_{pulley}$ is a radius of the pulley.
  We organize \equref{eq:basic-1} and \equref{eq:basic-2} with $K=R_{e}(D_{pulley}/(E_{gear}D_{gear}E_{motor}K_{t}))^{2}$, as shown below,
  \begin{align}
    \dot{c}_{1} = \frac{K}{C_{1}}f^{2} - \frac{c_{1}-c_{2}}{R_{1}C_{1}} \label{eq:basic-3}\\
    \dot{c}_{2} = \frac{c_{1}-c_{2}}{R_{1}C_{2}} - \frac{c_{2}-c_{a}}{R_{2}C_{2}} \label{eq:basic-4}
  \end{align}
  where $\dot{c}_{\{1, 2\}}$ represents $dc_{\{1, 2\}}/dt$.
  $c_{1}$ can be calculated by discretely repeating the recurrence relations of \equref{eq:basic-3} and \equref{eq:basic-4}.
  Also, in the usual case that $c_{2}$ can be obtained from a thermal sensor attached to motor housing, $c_{1}$ can be calculated just by repeating \equref{eq:basic-3}.

  However, the actual thermal parameters are different from the parameters obtained from datasheets.
  This is considered to be because of heat dissipation to the attached metal parts, error of the ambient temperature, deterioration or burnout of motors, etc.
  As one example, we compared a new motor and an old motor used for over half a year, whose rotation is inferior.
  We show the difference of the transition of $c_{2}$ when applying $f=200$ [N] over 90 seconds in \figref{figure:thermal-difference}.
  $c_{2}$ of the two motors were different by about 15 $^\circ$C in 90 seconds.
  Because $c_{1}$ is more sensitive than $c_{2}$, a larger difference must be generated.
  Thus, the thermal model should always be updated, and we propose the method below.
}%
{%
  本研究では, \cite{urata2008thermal}で述べられたものと同じ, \figref{figure:basic-model}の2抵抗モデルをモータの温度モデルとして使用する.
  なお, 本研究では温度の単位は$^\circ$C, 筋張力の単位は$N$, 筋長の単位は$mm$とする.
  モータコアとハウジングに熱容量$C_{1}, C_{2}$, モータコアとモータハウジング間, モータハウジングと外気温間に熱抵抗$R_{1}, R_{2}$を仮定する.
  このとき, モータコア温度$c_{1}$, モータハウジング温度$c_{2}$, 外気温$c_{a}$の間には以下のような関係が存在する.
  \begin{align}
    C_{1}\frac{dc_{1}}{dt} &= Q - \frac{c_{1}-c_{2}}{R_{1}} \label{eq:basic-1}\\
    C_{2}\frac{dc_{2}}{dt} &= \frac{c_{1}-c_{2}}{R_{1}} - \frac{c_{2}-c_{a}}{R_{2}} \label{eq:basic-2}\\
    Q &= R_{e}i^{2}
  \end{align}
  ここで, $R_{e}$は巻線電気抵抗, $i$は電流値である.
  加えて, 本研究の趣旨に沿うように, このモデルの熱量$Q$を電流ではなく筋張力で表したい場合は, 以下の式を用いて電流$i$を筋張力$f$に変換する.
  \begin{align}
    D_{pulley}f =E_{gear}D_{gear}E_{motor}K_{t}i
  \end{align}
  ここで, $K_{t}$はトルク定数, $E_{\{motor, gear\}}$はモータまたはギアの伝達効率, $D_{gear}$はギア比, $D_{pulley}$はプーリの半径を表す.
  よって, $K=R_{e}(D_{pulley}/(E_{gear}D_{gear}E_{motor}K_{t}))^{2}$として\equref{eq:basic-1}, \equref{eq:basic-2}を整理すると, 以下のようになる.
  \begin{align}
    \dot{c}_{1} = \frac{K}{C_{1}}f^{2} - \frac{c_{1}-c_{2}}{R_{1}C_{1}} \label{eq:basic-3}\\
    \dot{c}_{2} = \frac{c_{1}-c_{2}}{R_{1}C_{2}} - \frac{c_{2}-c_{a}}{R_{2}C_{2}} \label{eq:basic-4}
  \end{align}
  ここで, $\dot{c}_{\{1, 2\}}$は$dc_{\{1, 2\}}/dt$を表す.
  これら全てのパラメータはモータのデータシートから得ることができ, この漸化式\equref{eq:basic-3}, \equref{eq:basic-4}を離散的に繰り返すことで, モータコア温度$c_{1}$を得ることが出来る.
  またよくある構成として, モータのハウジングに温度センサを取り付けて$c_{2}$を得ることができる場合, \equref{eq:basic-3}のみを繰り返すことで$c_{1}$を得ることができる.

  しかし, 実際のパラメータはデータシートから得られるものとは大きく異なる.
  これは, モータに接続する金属等への放熱や外気温の違い, モータの経年劣化や故障など, 様々な要因が考えられる.
  その例として, 新品のモータと半年間用いた回りの悪いモータを比べた.
  一定の200 Nを90秒間加えた際のモータハウジングの温度上昇を\figref{figure:thermal-difference}に示す.
  温度上昇は90秒後で約15度程度異なっていることがわかる.
  モータコア温度はさらにセンシティブなため, より大きな差が生まれていると考えられる.
  よって, 温度モデルは常に学習し続けられていくべきであると考え, その方法を以下で提案する.
}%

\subsection{Proposed Thermal Model}
\switchlanguage%
{%
  We construct a model shown below, by setting the parameters in \equref{eq:basic-3} and \equref{eq:basic-4} as variables,
  \begin{align}
    \dot{c}_{1} = W_{1}\textrm{exp}(P_{1})f^{2} - \frac{c_{1}-c_{2}}{W_{2}\textrm{exp}(P_{2})} \label{eq:proposed-1}\\
    \dot{c}_{2} = \frac{c_{1}-c_{2}}{W_{3}\textrm{exp}(P_{3})} - \frac{c_{2}-W_{5}(1+P_{5})}{W_{4}\textrm{exp}(P_{4})} \label{eq:proposed-2}
  \end{align}
  where $W_{1}=K/C_{1}$, $W_{2}=R_{1}C_{1}$, $W_{3}=R_{1}C_{2}$, $W_{4}=R_{2}C_{2}$, and $W_{5}=c_{a}$.
  $P_{\{1, 2, 3, 4, 5\}}$ are parameters updated in this study, and $\textrm{exp}(\cdot)$ is an exponential function.
  The reason why the exponential function is applied to $P_{\{1, 2, 3, 4\}}$ is because  $W_{\{1, 2, 3, 4\}}\textrm{exp}(P_{\{1, 2, 3, 4\}})$ must be positive.
  When $P_{\{1, 2, 3, 4, 5\}}=0$, this model is equivalent with \equref{eq:basic-3} and \equref{eq:basic-4}.
    We can interpret $P_{1}$ as a coefficient to determine the quantity of heat from the current, $P_{2}$ as a time constant of heat escaping from the motor core to the motor housing, $P_{3}$ as a time constant of heat flowing in the motor housing from the motor core, $P_{4}$ as a time constant of heat escaping from the motor housing to the ambient, and $P_{5}$ as a coefficient expressing the ambient temperature ($W_{5}(1+P_{5})$ expresses the updated ambient temperature).

  We can simply express this model by using the function of $h_{1}$ and $h_{2}$, as shown below.
  \begin{align}
    \dot{c}_{1} = h_{1}(f, c_{1}, c_{2}) \label{eq:proposed-3}\\
    \dot{c}_{2} = h_{2}(c_{1}, c_{2}) \label{eq:proposed-4}
  \end{align}
  In this study, we update $P_{\{1, 2, 3, 4, 5\}}$ to match the actual parameters of the robot online for the accurate thermal estimation and control using \equref{eq:proposed-3} and \equref{eq:proposed-4}.
}%
{%
  本研究では, \equref{eq:basic-3}, \equref{eq:basic-4}におけるパラメータを変数としておき, 以下のようなモデルを作成する.
  \begin{align}
    \dot{c}_{1} = W_{1}\textrm{exp}(P_{1})f^{2} - \frac{c_{1}-c_{2}}{W_{2}\textrm{exp}(P_{2})} \label{eq:proposed-1}\\
    \dot{c}_{2} = \frac{c_{1}-c_{2}}{W_{3}\textrm{exp}(P_{3})} - \frac{c_{2}-W_{5}(1+P_{5})}{W_{4}\textrm{exp}(P_{4})} \label{eq:proposed-2}
  \end{align}
  ここで, $W_{1}=K/C_{1}$, $W_{2}=R_{1}C_{1}$, $W_{3}=R_{1}C_{2}$, $W_{4}=R_{2}C_{2}$, $W_{5}=c_{a}$とおいた.
  $P_{\{1, 2, 3, 4, 5\}}$は本モデルで学習されるパラメータであり, $\textrm{exp}(\cdot)$は指数関数である.
  $P_{\{1, 2, 3, 4\}}$を指数関数に通したのは, $W_{\{1, 2, 3, 4\}}\textrm{exp}(P_{\{1, 2, 3, 4\}})$を必ず正とするためである.
  $P_{\{1, 2, 3, 4, 5\}}=0$のとき, このモデルは\equref{eq:basic-3}, \equref{eq:basic-4}と一致する.
  これらのパラメータは, $P_{1}$は電流からの熱量を決める係数, $P_{2}$はモータコアからモータハウジングに逃げる熱の時定数, $P_{3}$はモータコアからモータハウジングに流入する熱の時定数, $P_{4}$はモータハウジングから外部に逃げる熱の時定数, $P_{5}$は外気温度を表す係数, と解釈が可能である.

  このモデルを関数$h_{1}, h_{2}$を使って簡易に表すと, 以下のようになる.
  \begin{align}
    \dot{c}_{1} = h_{1}(f, c_{1}, c_{2}) \label{eq:proposed-3}\\
    \dot{c}_{2} = h_{2}(c_{1}, c_{2}) \label{eq:proposed-4}
  \end{align}
  本研究では, $P_{\{1, 2, 3, 4, 5\}}$をパラメータとして, \equref{eq:proposed-3}, \equref{eq:proposed-4}を使って温度推定, 温度制限制御を行っていくと同時に, これらパラメータをオンラインで実機に合うように更新していく.
}%

\subsection{Thermal Estimator: Estimation of Motor Core Temperature} \label{subsec:thermal-estimator}
\switchlanguage%
{%
  In this study, as a common configuration of motors, we assume that a thermal sensor is attached to the motor housing.
  In this case, a method to estimate $c_{1}$ is simple.
  We determine an update interval $\Delta{t}_{est}$ and estimate $c_{1}$ as below,
  \begin{align}
    c_{1, t+1} = c_{1, t} + h_{1}(f_{t}, c_{1, t}, c_{2, t})\Delta{t}_{est} \label{eq:estimate}
  \end{align}
  where $\{c_{1}, c_{2}, f\}_{t}$ is $\{c_{1}, c_{2}, f\}$ at the current time step $t$.

  In this study, we set $\Delta{t}_{est}=0.02$ [sec].
}%
{%
  本研究では, 一般的な構成として, モータのハウジング部に温度センサが取り付けられたような構成を考える.
  この場合, モータコア温度推定の方法は非常に単純である.
  温度推定の時間間隔$\Delta{t}_{est}$を決め, 以下のように$c_{1}$を推定する.
  \begin{align}
    c_{1, t+1} = c_{1, t} + h_{1}(f_{t}, c_{1, t}, c_{2, t})\Delta{t}_{est} \label{eq:estimate}
  \end{align}
  ここで, $\{c_{1}, c_{2}, f\}_{t}$は現在のタイムステップ$t$における$\{c_{1}, c_{2}, f\}$を表す.

  本研究では, $\Delta{t}_{est}=0.02$ [sec]としている.
}%

\subsection{Online Learning of Thermal Model} \label{subsec:online-learning}
\switchlanguage%
{%
  By using the obtained sensor data of $c_{2}$ and $f$, we update $P_{\{1, 2, 3, 4, 5\}}$.

  First, we accumulate sensor data.
  We set the interval of data accumulation $\Delta{t}_{data}$, the number of sequences of one batch $N_{seq}$, and the batch size $N_{batch}$.
  We accumulate $\{c_{1}, c_{2}, f\}^{data}$ which is $N_{seq}$ consecutive data of $\{c_{1}, c_{2}, f\}$ at intervals of $\Delta{t}_{data}$.
  Because $c_{1}$ cannot be directly obtained, we accumulate its estimated value explained in \secref{subsec:thermal-estimator}.
  We accumulate the batch of $N_{seq}$ number of $\{c_{1}, c_{2}, f\}^{data}$, and begin to update the thermal model when the number of the batches exceeds $N_{batch}$.
  After finishing the update of the thermal model using $N_{batch}$ number of data batches, we remove the first of $N_{batch}$ batches, and update the model when the number of batches exceeds $N_{batch}$ again.
  These procedures are repeated online.

  Next, we will explain the details of the online model update.
  Although we use only \equref{eq:proposed-3} for the estimation of $c_{1}$, we can also estimate $c_{2}$ by \equref{eq:proposed-4} at the same time.
  By comparing the estimated $c_{2}$ and the actual obtained $c_{2}$, we can update $P_{\{1, 2, 3, 4, 5\}}$.
  Among the accumulated data sequence of $\{c_{1}, c_{2}, f\}^{data}_{[k, k+N_{seq}-1]}$, first we pick out $c^{data}_{1, k}$, $c^{data}_{2, k}$, and $f^{data}_{[k, k+N_{seq}-1]}$.
  As $c_{1, k} = c^{data}_{1, k}$, $c_{2, k} = c^{data}_{2, k}$, and $f_{[k, k+N_{seq}-1]}=f^{data}_{[k, k+N_{seq}-1]}$, we repeat the equation below $N_{seq}-1$ times.
  \begin{align}
    c_{1, k+1} = c_{1, k} + h_{1}(f_{k}, c_{1, k}, c_{2, k})\Delta{t}_{data} \label{eq:update-1}\\
    c_{2, k+1} = c_{2, k} + h_{2}(c_{1, k}, c_{2, k})\Delta{t}_{data} \label{eq:update-2}
  \end{align}
  Then, we can obtain $c_{1, [k+1, k+N_{seq}-1]}$ and $c_{2, [k+1, k+N_{seq}-1]}$.
  We compare the accumulated data of $c^{data}_{2, [k+1, k+N_{seq}-1]}$ and the data of $c_{2, [k+1, k+N_{seq}-1]}$ estimated using the current parameters, by the equation below,
  \begin{align}
    L_{update} = \textrm{MSE}(c_{2, [k+1, k+N_{seq}-1]}, c^{data}_{2, [k+1, k+N_{seq}-1]})
  \end{align}
  where $\textrm{MSE}$ represents mean squared error.
  Then, we update $P_{\{1, 2, 3, 4, 5\}}$ from this loss function by backpropagation through time \cite{rumelhart1986bptt}, as shown below,
  \begin{align}
    P_{\{1, 2, 3, 4, 5\}} \gets P_{\{1, 2, 3, 4, 5\}} - \alpha \frac{\partial{L}_{update}}{\partial{P_{\{1, 2, 3, 4, 5\}}}}
  \end{align}
  where $\alpha$ is a learning rate.
  This is equivalent to updating weights of a neural network that represent parameters of the thermal model by a gradient descent method.
  Here, these parameters are actually updated using the average of the gradient calculated from the $N_{batch}$ data.
  Also, we set the maximum norm of the gradient as $D_{clip}$ by a gradient clipping method \cite{pascanu2013clipping}.

  The problem of this method is that $c^{data}_{1, k}$ is not the actual sensor data but the estimated value.
  However, the change in $c_{2}$ is not very sensitive to the initial value of $c_{1}$, and from the subsequent experiments, we can verify that the parameters can be updated correctly.

  In this study, we set $\Delta{t}_{data}=1.0$ [sec], $N_{seq}=30$, $N_{batch}=10$, $\alpha=0.02$, and $D_{clip}=5.0$.
  These parameters are set from preliminary experiments to quickly converge the online update and not to exceed the time limit.
}%
{%
  モータのハウジング部に取り付けられた温度センサの値$c_{2}$と筋張力$f$を教師として, 温度モデルのパラメータ$P_{\{1, 2, 3, 4, 5\}}$を更新していく.

  まず, 教師となるデータを蓄積する.
  データの蓄積間隔$\Delta{t}_{data}$, データ列数$N_{seq}$, バッチ数$N_{batch}$を設定する.
  $\Delta{t}_{data}$の間隔で連続する$N_{seq}$個の$\{c_{1}, c_{2}, f\}$のデータ$\{c_{1}, c_{2}, f\}^{data}$を蓄積する.
  ここで, $c_{1}$は直接得られないため, 前節で得られた推定値を蓄積している.
  この$N_{seq}$個のデータ列を貯めていき, それらが$N_{batch}$個得られたところで学習を実行し始める.
  これら$N_{batch}$個のデータ集合を使って学習が終わったら, その内一番最初のデータを削除し, また$N_{seq}$個のデータ列が集まりデータ集合が$N_{batch}$個になったところでまた学習を実行することをオンラインで繰り返す.

  次に, オンライン学習について説明する.
  モータコア温度推定には\equref{eq:proposed-3}のみしか用いなかったが, \equref{eq:proposed-4}によって, 同時に$c_{2}$を推定することが可能である.
  そこで, 推定された$c_{2}$と, 実際に得られた$c_{2}$を比較することで, パラメータ$P_{\{1, 2, 3, 4, 5\}}$を更新していくことになる.
  蓄積されたデータ列$\{c_{1}, c_{2}, f\}^{data}_{[k, k+N_{seq}-1]}$のうち, はじめに$c^{data}_{1, k}, c^{data}_{2, k}, f^{data}_{[k, k+N_{seq}-1]}$を取り出す.
  その後, $c_{1, k} = c^{data}_{1, k}, c_{2, k} = c^{data}_{2, k}, f_{[k, k+N_{seq}-1]}=f^{data}_{[k, k+N_{seq}-1]}$として, 以下の漸化式を$N_{seq}-1$回繰り返す.
  \begin{align}
    c_{1, k+1} = c_{1, k} + h_{1}(f_{k}, c_{1, k}, c_{2, k})\Delta{t}_{data} \label{eq:update-1}\\
    c_{2, k+1} = c_{2, k} + h_{2}(c_{1, k}, c_{2, k})\Delta{t}_{data} \label{eq:update-2}
  \end{align}
  すると, $c_{1, [k+1, k+N_{seq}-1]}, c_{2, [k+1, k+N_{seq}-1]}$が得られる.
  ここで, 実機からデータ列として得られた$c^{data}_{2, [k+1, k+N_{seq}-1]}$と現在のパラメータによって推定された$c_{2, [k+1, k+N_{seq}-1]}$を以下の式で比較する.
  \begin{align}
    L_{update} = \textrm{MSE}(c_{2, [k+1, k+N_{seq}-1]}, c^{data}_{2, [k+1, k+N_{seq}-1]})
  \end{align}
  ここで, $\textrm{MSE}$はMean Squared Errorを表す.
  そして, この誤差からパラメータ$P_{\{1, 2, 3, 4, 5\}}$を通時的誤差逆伝播法\cite{rumelhart1986bptt}を用いて以下のように更新する.
  \begin{align}
    P_{\{1, 2, 3, 4, 5\}} \gets P_{\{1, 2, 3, 4, 5\}} - \alpha \frac{\partial{L}_{update}}{\partial{P_{\{1, 2, 3, 4, 5\}}}}
  \end{align}
  ここで, $\alpha$は学習率である.
  これは言ってみれば, ニューラルネットワークのパラメータを直接温度推定のパラメータとして, 最急降下法によって更新していることに相当する.
  このとき, 実際には$N_{batch}$個のデータから計算された勾配の平均を持ってパラメータを更新する.
  また, このときGradient Clipping \cite{pascanu2013clipping}を用いて, 勾配のノルムの最大値を$D_{clip}$に設定している.

  この手法の問題点として, $c^{data}_{1, k}$が実測値ではなく推定値であることが挙げられる.
  しかし, $c_{2}$の変化は$c_{1}$の初期値に対してそこまで敏感ではなく, 後の実験からも本手法により正確にパラメータが更新できることがわかる.

  本研究では, $\Delta{t}_{data}=1.0$ [sec], $N_{seq}=30$, $N_{batch}=10$, $\alpha=0.02$, $D_{clip}=5.0$としている.
}%

\subsection{Anomaly Detection of Motors} \label{subsec:anomaly-detection}
\switchlanguage%
{%
  In this study, because the updated $P_{\{1, 2, 3, 4, 5\}}$ have physical meanings, we can conduct anomaly detection using the change in these parameters.
  This is the difference from a neural network whose parameters are difficult to interpret.
  We regard the current situation as anomaly simply when the value $g$ shown below exceeds a threshold $D_{detect}$,
  \begin{align}
    g = \textrm{RMSE}([P_{1}, P_{2}, P_{3}, P_{4}]^T, [P^{init}_{1}, P^{init}_{2}, P^{init}_{3}, P^{init}_{4}]^T)
  \end{align}
  where $P^{init}_{\{1, 2, 3, 4\}}$ is $P_{\{1, 2, 3, 4\}}$ at the start of the experiment, and $\textrm{RMSE}$ represents root mean squared error.
  Because $P_{5}$ is a parameter of the ambient temperature and can always change, we do not use it for $g$.
  Also, when executing online learning for the first time, we start with $P_{\{1, 2, 3, 4\}}=0$, and $P_{\{1, 2, 3, 4\}}$ can change greatly.
  Therefore, anomaly detection should be executed after the parameters are firmly updated once.
  Also, one of the simplest methods of anomaly detection is displaying and monitoring these four parameters at all times.

  In this study, we set $D_{detect}=1.0$.
}%
{%
  本研究では, オンラインで学習されたパラメータ$P_{\{1, 2, 3, 4, 5\}}$が物理的な意味を持つため, この変化を用いて異常検知を行うことができる.
  これは, 物理的な意味として解釈が難しい重みを学習するニューラルネットワークとの大きな違いである.
  異常検知は単純に, 以下の値$g$が閾値$D_{detect}$を超えたら異常と見なす.
  \begin{align}
    g = \textrm{RMSE}([P_{1}, P_{2}, P_{3}, P_{4}]^T, [P^{init}_{1}, P^{init}_{2}, P^{init}_{3}, P^{init}_{4}]^T)
  \end{align}
  ここで, $P^{init}_{\{1, 2, 3, 4\}}$は動作を開始したときの$P_{\{1, 2, 3, 4\}}$, $\textrm{RMSE}$はRoot Mean Squared Errorを表す.
  $P_{5}$は外気温のパラメータであり常に変わる可能性があるため, これは$g$から抜いている.
  初めてオンライン学習を実行する際は$P_{\{1, 2, 3, 4\}}=0$から始めるため, $P_{\{1, 2, 3, 4\}}$は大きく変化する可能性がある.
  そのため, 一度パラメータがしっかりと学習されてから, 異常検知は行うと良い.
  また, もう一つの最も単純な異常検知の方法は, この4つのパラメータを表示して監視することである.

  本研究では, $D_{detect}=1.0$としている.
}%

\subsection{Thermal Controller: Control of Motor Core Temperature} \label{subsec:thermal-controller}
\switchlanguage%
{%
  The thermal controller is a control to calculate the smooth sequence of maximum muscle tension $f^{limit}$ to rapidly achieve the maximum motor core temperature $c^{max}_{1}$ and to restrict motor output by this value.
  In this section, we will only explain the optimization part.
  We set the time interval $\Delta{t}_{control}$ and the number of sequences to consider $N_{control}$.
  Then, we optimize the sequence of $f$ during $N_{control}\Delta{t}_{control}$ seconds to achieve $c^{max}_{1}$ as rapidly as possible.

  First, we represent the current estimated $c_{1}$ as $c^{current}_{1, t}$, and $c_{2}$ obtained from the thermal sensor as $c^{current}_{2, t}$.
  Also, we determine the sequence of $f$ before optimization $f^{limit}_{[k, k+N_{control}-1]}$.
  By setting $c_{1, k} = c^{current}_{1, k}$, $c_{2, k} = c^{current}_{2, k}$, and $f_{[k, k+N_{control}-1]}=f^{limit}_{[k, k+N_{control}-1]}$, we repeat the recurrence relations below $N_{control}-1$ times.
  \begin{align}
    c_{1, k+1} = c_{1, k} + h_{1}(f_{k}, c_{1, k}, c_{2, k})\Delta{t}_{control} \label{eq:control-1}\\
    c_{2, k+1} = c_{2, k} + h_{2}(c_{1, k}, c_{2, k})\Delta{t}_{control} \label{eq:control-2}
  \end{align}
  Then, we can obtain $c_{1, [k+1, k+N_{control}-1]}$ and $c_{2, [k+1, k+N_{control}-1]}$.
  We calculate the loss between $c^{max}_{1, [k+1, k+N_{control}-1]}$, which is a vector in which $N_{control}$ numbers of $c^{max}_{1}$ are arranged, and the estimated $c_{1, [k+1, k+N_{control}-{1}]}$ as below,
  \begin{align}
    L_{control} &= \textrm{MSE}(c_{1, [k+1, k+N_{control}-1]}, c^{max}_{1, [k+1, k+N_{control}-1]})\nonumber\\
    &\;\;\;\;\;+W_{control}\textrm{MSE}(\bm{0}, f^{limit}_{[k, k+N_{control}-1]})
  \end{align}
  where $W_{control}$ is a constant weight.
  By adding the minimization term of $f^{limit}$, the transition of $f^{limit}$ optimized from $L_{control}$ becomes smooth, and the stability of optimization increases.
  Although the term minimizing the difference of $f^{limit}$ at adjacent time steps is suitable for the purpose of this study, we found that the term destabilizes the optimization from preliminary experiments.
  By using the loss of $L_{control}$, $f^{limit}_{[k, k+N_{control}-1]}$ is updated by backpropagation through time \cite{rumelhart1986bptt}, as shown below,
  \begin{align}
    f^{limit}_{[k, k+N_{control}-1]} \gets f^{limit}_{[k, k+N_{control}-1]} - \beta \frac{\partial{L}_{control}}{\partial f^{limit}_{[k, k+N_{control}-1]}}
  \end{align}
  where $\beta$ is a learning rate.

  In actuality, we determine the maximum and minimum value of muscle tension $f^{\{min, max\}}$ for safety, and restrict $f^{limit}$.
  Also, the initial value of $f^{limit}_{[k, k+N_{control}-1]}$ is determined as $[f^{limit, T}_{[k, k+N_{control}-2]}, f^{limit}_{k+N_{control}-2}]^{T}$ by combining the shifted value of the previously optimized sequence of $f^{limit}_{[k-1, k+N_{control}-2]}$ and its last term.
  When it is the first time to optimize and there is no previously optimized value, we use a value in which is a vector that $N_{control}$ numbers of $f^{max}$ are arranged.

  In this study, we set $\Delta{t}_{control}=1.0$ [sec], $c^{max}_{1}=80$ [$^\circ$C], $N_{control}=30$, $W_{control}=0.001$, $\beta=30$ [N], $f^{min}=10$ [N], and $f^{max}=300$ [N].
  From preliminary experiments, $\Delta{t}_{control}$ and $N_{control}$ are set not to exceed the time limit, $c^{max}_{1}$ and $f^{\{min, max\}}$ are set within the range of ordinary use, and $W_{control}$ and $\beta$ are set to quickly stabilize the calculation of $f^{limit}$.
}%
{%
  モータコア温度制限制御は, モータコア温度$c_{1}$が設定した最大値$c^{max}_{1}$に達するような滑らかな筋張力$f^{limit}$の時系列を求め, その値を持って筋張力を制限する制御である.
  これは, ある時間間隔$\Delta{t}_{control}$とそれを展開する回数$N_{control}$を決め, $\Delta{t}_{control}N_{control}$秒間でなるべく速く$c_{1}$が$c^{max}_{1}$を達成するための滑らかな$f$の時系列を最適化することに相当する.
  まず, 現在推定されているモータコア温度を$c^{current}_{1, t}$, 温度センサから得られたハウジング温度を$c^{current}_{2, t}$とする.
  また, 最適化する前の$f$の時系列$f^{limit}_{[k, k+N_{control}-1]}$を決定する.
  これらから, $c_{1, k} = c^{current}_{1, k}, c_{2, k} = c^{current}_{2, k}, f_{[k, k+N_{control}-1]}=f^{limit}_{[k, k+N_{control}-1]}$として, 以下の漸化式を$N_{control}-1$回繰り返す.
  \begin{align}
    c_{1, k+1} = c_{1, k} + h_{1}(f_{k}, c_{1, k}, c_{2, k})\Delta{t}_{control} \label{eq:control-1}\\
    c_{2, k+1} = c_{2, k} + h_{2}(c_{1, k}, c_{2, k})\Delta{t}_{control} \label{eq:control-2}
  \end{align}
  すると, $c_{1, [k+1, k+N_{control}-1]}, c_{2, [k+1, k+N_{control}-1]}$が得られる.
  ここで, $c_{1}$の最大値$c^{max}_{1}$を$N_{control}$個並べた$c^{max}_{1, [k+1, k+N_{control}-1]}$と, 推定された$c_{1, [k+1, k+N_{control}-{1}]}$の誤差を以下の式で比較する.
  \begin{align}
    L_{control} &= \textrm{MSE}(c_{1, [k+1, k+N_{control}-1]}, c^{max}_{1, [k+1, k+N_{control}-1]})\nonumber\\
    &\;\;\;\;\;+W_{control}\textrm{MSE}(\bm{0}, f^{limit}_{[k, k+N_{control}-1]})
  \end{align}
  ここで, $W_{control}$は右辺第一項と, $f^{limit}$を最小化するような右辺第二項をバランシングするための重みである.
  $f^{limit}$の最小化項を加えることで, 以降の$L_{control}$を用いて最適化される$f^{limit}$の遷移が滑らかになり, 最適化の際の安定性が増す.
  隣り合うタイムステップ同士の$f^{limit}$を最小化する方が本研究の趣旨には合うが, 実験からその項を入れると不安定になることがわかっている.
  この誤差から$f^{limit}_{[k, k+N_{control}-1]}$を通時的誤差逆伝播法\cite{rumelhart1986bptt}を用いて以下のように更新する.
  \begin{align}
    f^{limit}_{[k, k+N_{control}-1]} \gets f^{limit}_{[k, k+N_{control}-1]} - \beta \frac{\partial{L}_{control}}{\partial f^{limit}_{[k, k+N_{control}-1]}}
  \end{align}
  ここで, $\beta$は学習率である.

  実際には安全対策として筋張力の最大値と最小値$f^{\{min, max\}}$を決め, それよりも$f^{limit}$が大きくならないように制限している.
  また, 最初に説明した$f^{limit}_{[k, k+N_{control}-1]}$の初期値は, 前ステップで最適化された値$f^{limit}_{[k-1, k+N_{control}-2]}$をshiftして最終項を複製した$[f^{limit, T}_{[k, k+N_{control}-2]}, f^{limit}_{k+N_{control}-2}]^{T}$としている.
  この最適化が初めて, つまり前ステップで最適化された値がない場合は$f^{max}$が$N_{control}$個並んだものを用いる.

  本研究では, $\Delta{t}_{control}=1.0$ [sec], $c^{max}_{1}=80$ [$^\circ$C], $N_{control}=30$, $W_{control}=0.001$, $\beta=30$ [N], $f^{min}=10$ [N], $f^{max}=300$ [N]としている.
}%
\subsection{Muscle Tension Limiter} \label{subsec:tension-limiter}
\switchlanguage%
{%
  The muscle tension limiter restricts $f$ by the maximum value of $f^{limit}_{k}$ calculated by \secref{subsec:thermal-controller}.
  Although we can set the maximum muscle tension easily when using tension-based control \cite{jantsch2012computed}, when using length-based control, we elongate the muscle length and send $l^{ref} + \Delta{l}_{k}$ to the robot as below, as the muscle does not vibrate,
  \begin{align}
    &if\;\;f > f^{limit}_{k} \nonumber\\
    &\;\;\;\;\;\;\;\;\;\;\;\;\Delta{l}_{k} = \Delta{l}_{k-1} + min(D_{gain}d-\Delta{l}_{k-1}, \Delta{l}_{plus}d)\\
    &else \nonumber\\
    &\;\;\;\;\;\;\;\;\;\;\;\;\Delta{l}_{k} = \Delta{l}_{k-1} + max(0-\Delta{l}_{k-1}, -\Delta{l}_{minus}d)\\
    &d = |f-f^{limit}_{k}|&
  \end{align}
  where $|\cdot|$ is an absolute value, $\Delta{l}_{\{minus, plus\}}$ is a coefficient to determine the change in muscle length at one time step in the negative or positive direction, and $D_{gain}$ is a coefficient to determine the maximum elongation value.
  Thus, while restricting the change in muscle length by $\Delta{l}_{minus}d$ and $\Delta{l}_{plus}d$, muscle length is elongated or pulled as $f$ does not exceed $f^{limit}_{k}$.
  By using this control, even when an excessive muscle length is sent to the robot, the muscle actuator elongates automatically so that $c_{1}$ does not exceed $c^{max}_{1}$.

  In this study, we set $\Delta{l}_{minus}=0.001$, $\Delta{l}_{plus}=0.003$, and $D_{gain}=2.0$, and this control is executed every $8$ msec.
  These parameters are set from preliminary experiments to limit muscle tension quickly and to not vibrate.
}%
{%
  得られた$f^{limit}_{k}$を現在の最大筋張力に設定し, 筋張力制限をかける.
  筋張力制御\cite{jantsch2012computed}であれば設定する筋張力最大値をこの値に設定するだけであるが, 筋長制御の場合は以下のように, 振動しないように筋長を緩め$l^{ref} + \Delta{l}_{k}$をロボットに送る.
  \begin{align}
    &if\;\;f > f^{limit}_{k} \nonumber\\
    &\;\;\;\;\;\;\;\;\;\;\;\;\Delta{l}_{k} = \Delta{l}_{k-1} + min(D_{gain}d-\Delta{l}_{k-1}, \Delta{l}_{plus}d)\\
    &else \nonumber\\
    &\;\;\;\;\;\;\;\;\;\;\;\;\Delta{l}_{k} = \Delta{l}_{k-1} + max(0-\Delta{l}_{k-1}, -\Delta{l}_{minus}d)\\
    &d = |f-f^{limit}_{k}|&
  \end{align}
  ここで, $|\cdot|$は絶対値, $l^{ref}$は実機に送る筋長指令, $\Delta{l}_{k}$はタイムステップ$k$における弛緩度, $\Delta{l}_{\{minus, plus\}}$はマイナス方向またはプラス方向に対する一ステップの筋長変化量を決める係数, $D_{gain}$は最大弛緩量を決める係数である.
  つまり, $\Delta{l}_{minus}d$, $\Delta{l}_{plus}d$で制限をかけながら, 筋張力が最大値を越えないように筋を弛緩・緊張させている.
  この制御により, 無理な筋長が送られたり力がかかっても, モータコア温度が$c^{max}_{1}$を越えないように, 勝手に筋が弛緩するようになる.

  本研究では, $\Delta{l}_{minus}=0.001$, $\Delta{l}_{plus}=0.003$, $D_{gain}=2.0$とし, 本制御は$8$ msec周期で行う.
}%

\begin{figure}[t]
  \centering
  \includegraphics[width=0.9\columnwidth]{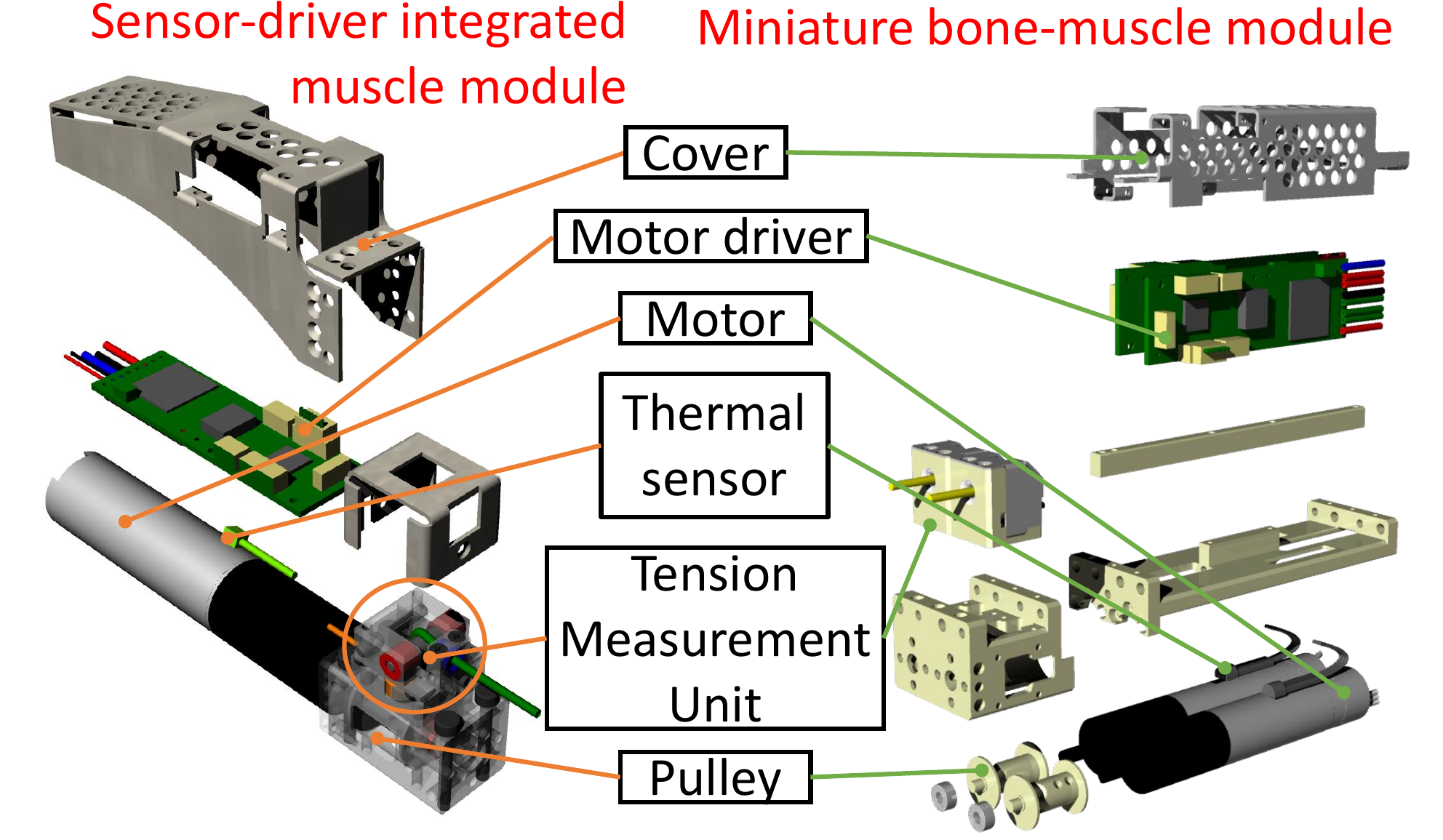}
  \caption{Muscle actuator configuration in Musashi: sensor-driver integrated muscle module \cite{asano2015sensordriver} and miniature bone-muscle module \cite{kawaharazuka2017forearm}.}
  \label{figure:muscle-actuator}
  \vspace{-3.0ex}
\end{figure}

\section{Experiments} \label{sec:experiment}
\switchlanguage%
{%
  We conducted (A) a simulation and (B) actual motor experiments using one muscle actuator, as well as (C) using multiple muscle actuators of the musculoskeletal humanoid.

  We show the configuration of muscle actuators of Musashi \cite{kawaharazuka2019musashi} used in this study: sensor-driver integrated muscle module \cite{asano2015sensordriver} and miniature bone-muscle module \cite{kawaharazuka2017forearm}, in \figref{figure:muscle-actuator}.
  Maxon brushless DC motor of EC-4pole 22 90W with 29:1 gear ratio is used for the former and EC 16 60W with 128:1 gear ratio is used for the latter.
  The former is used for the shoulder and elbow, and the latter is used for the wrist and fingers.
  As stated so far, a thermal sensor is attached to the motor housing.
  The wire of the muscle is wound by a pulley, and it comes out from the tension measurement unit, which can measure $f$.
  A motor driver, which can do current control, is attached to the motor, with a cover.

  We describe the default parameters of these muscle actuators.
  Regarding EC-4pole 22 90W with 29:1 (sensor-driver integrated muscle module), $C_{1}= 2.10$ [J/K], $C_{2}=29.0$ [J/K], $R_{1}=1.20$ [K/W], $R_{2}=10.3$ [K/W], and $K=2.97E-4$ [J/N$^2$s].
  Regarding EC 16 60W with 128:1 (miniature bone-muscle module), $C_{1}=1.19$ [J/K], $C_{2}=12.2$ [J/K], $R_{1}=4.30$ [K/W], $R_{2}=39.5$ [K/W], and $K=4.50E-5$ [J/N$^2$s].
  All parameters can be obtained from the datasheet of motors.
  Only $C_{\{1, 2\}}$ cannot be directly obtained, but these parameters are approximated by $T_{\{1, 2\}}/R_{\{1, 2\}}$ ($T_{\{1, 2\}}$ are thermal time constants of motor core and housing).
  Also, we set $c_{a}=30$ [$^\circ$C] as a default value.
}%
{%
  一つの筋アクチュエータにおけるシミュレーション, 実機実験を行い, 最後に筋骨格ヒューマノイドの複数の筋における実機実験を行う.

  本研究で用いるMusashi\cite{kawaharazuka2019musashi}の筋アクチュエータ\cite{asano2015sensordriver, kawaharazuka2017forearm}の構成を\figref{figure:muscle-actuator}に示す.
  筋のモータはMaxonのbrushless DC motorである, EC-4pole 22 90W with 29:1 gearとEC 16 60W with 128:1を用いており, 肩・肘に用いられるのは前者, 手首・指に用いられるのは後者である.
  これまで述べたように, モータハウジングに温度センサを有している.
  また, モータの先についたプーリに筋が巻かれており, 筋張力を測定可能な筋張力測定ユニットから筋が出て行く.
  また, 電流制御可能なモータドライバがカバーで覆われモータ上部に接続されている.

  ここで, デフォルトのパラメータについて記載する.
  EC-4pole 22 90W with 29:1に関しては, $C_{1}= 2.10$ [J/K], $C_{2}=29.0$ [J/K], $R_{1}=1.20$ [K/W], $R_{2}=10.3$ [K/W], $K=2.97E-4$ [J/N$^2$s]である.
  また, EC 16 60W with 128:1に関しては, $C_{1}=1.19$ [J/K], $C_{2}=12.2$ [J/K], $R_{1}=4.30$ [K/W], $R_{2}=39.5$ [K/W], $K=4.50E-5$ [J/N$^2$s]である.
  $c_{a}$については, 本研究では30 $^\circ$Cとしている.
}%

\begin{figure}[t]
  \centering
  \includegraphics[width=1.0\columnwidth]{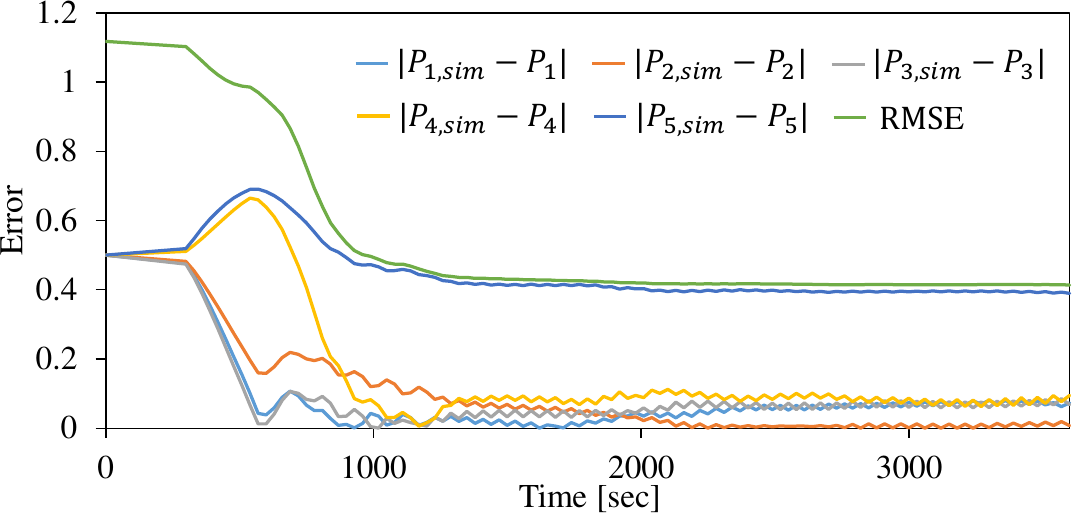}
  \caption{Transition of errors in thermal model parameters between simulator and current model during online learning.}
  \label{figure:simulation-learning}
  \vspace{-3.0ex}
\end{figure}

\begin{figure}[t]
  \centering
  \includegraphics[width=1.0\columnwidth]{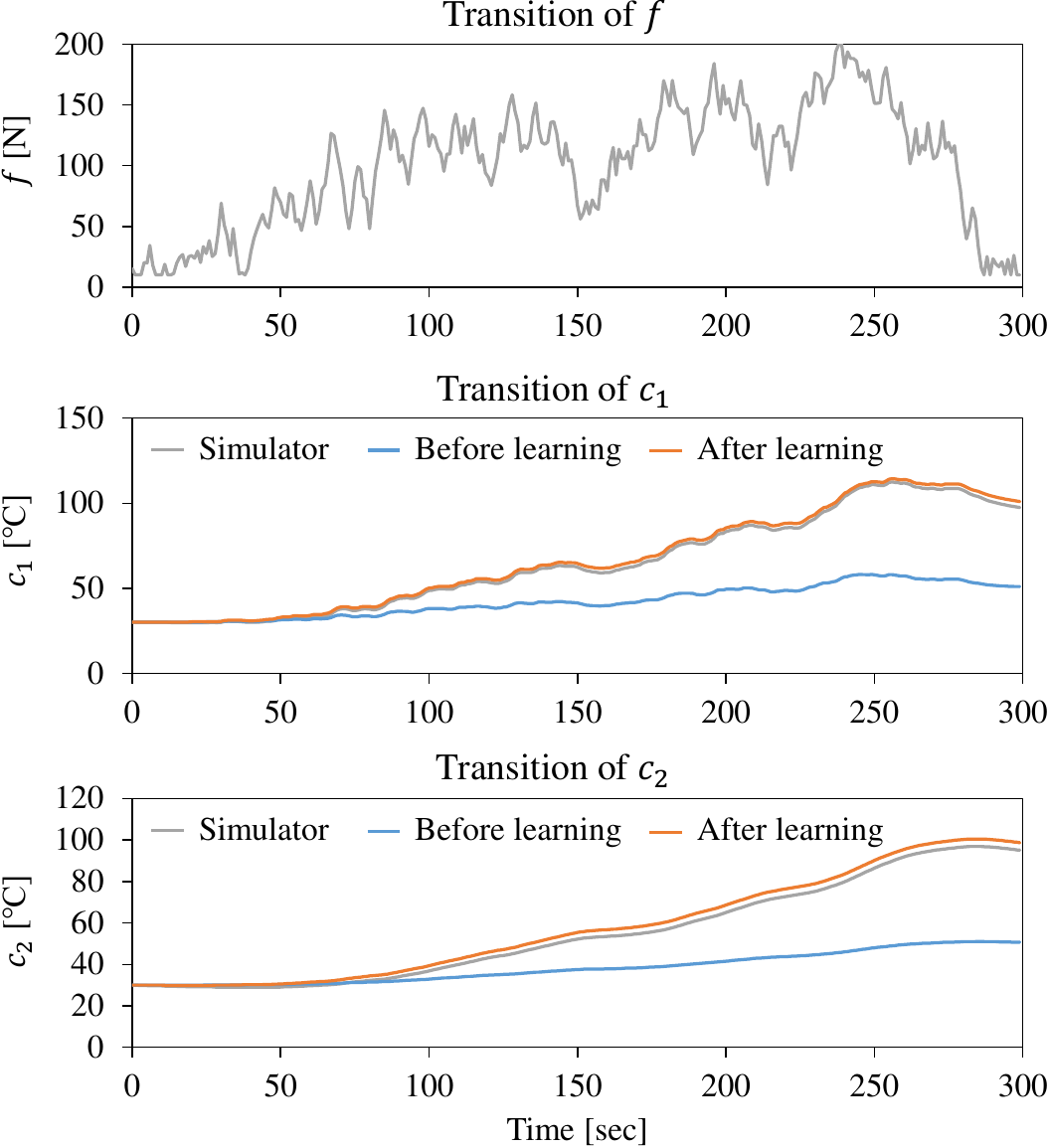}
  \caption{Transition of $(f, c_{1}, c_{2})$ of the simulator and the estimated $c_{1}$ and $c_{2}$ using the model before and after online learning.}
  \label{figure:simulation-evaluation}
  \vspace{-1.0ex}
\end{figure}

\subsection{Simulation Experiment of One Muscle Actuator}
\subsubsection{Online Learning} \label{subsubsec:simulation-learning}
\switchlanguage%
{%
  We used a simulator of EC-4pole 22 90W with modified parameters of $P_{1, sim}=0.5, P_{2, sim}=0.5, P_{3, sim}=-0.5, P_{4, sim}=-0.5, P_{5, sim}=0.5$.
  We set both the initial $c_{1}$ and $c_{2}$ as 30 $^\circ$C.
  $f$ was updated by the equation of $f \gets f+\textrm{Rand}(-50.0, 50.0)$ every $\Delta{t}_{data}$ seconds ($\textrm{Rand}(a, b)$ is a random value in $[a, b]$, and $f$ is limited within $[10, 200]$ [N]), and was sent to the simulator.
  Only $c_{2}$ and $f$ can be observed from the simulator.
  There is a thermal model corresponding to the simulator, with default parameters of $P_{\{1, 2, 3, 4, 5\}}=0$.
  By using this model, $c_{1}$ is estimated from the observed $c_{2}$ and $f$ (the initial value of $c_{1}$ is 30 $^\circ$C), and online learning of the current model is executed using the obtained $(c_{1}, c_{2}, f)$.
  This procedure makes the parameters of the current model closer to those of the simulator.
  We show the absolute errors of parameters between the simulator and current model $|P_{\{1, 2, 3, 4, 5\}, sim}-P_{\{1, 2, 3, 4, 5\}}|$ and the transition of RMSE of these five parameters over 3600 seconds in \figref{figure:simulation-learning}.
  We can see that each parameter of the current model became closer to that of the simulator, and the transition of errors converged in about 1200 seconds.
  Although the error of $P_{5}$ did not decrease to around 0, this is considered to be because the transition of $c_{2}$, which is effective in updating $P_{5}$, could not be obtained.

  Also, after executing online learning over 3600 seconds, we sent $f$ as stated above again to the simulator, the model before online learning, and the model after online learning.
  The result before online learning is the same with the result of the previous study \cite{urata2008thermal}.
  We show the transition of $(f, c_{1}, c_{2})$ in \figref{figure:simulation-evaluation}.
  The thermal estimation of $c_{1}$ and $c_{2}$ after online learning became closer to that of the simulator compared with before online learning.
  Although we used only $c_{2}$ as a teaching signal, $c_{1}$ also became closer to that of the simulator, and so this method is effective.
}%
{%
  まずオンライン学習について実験を行う.
  アクチュエータのパラメータはEC-4pole 22 90W with 29:1を$P_{1, sim}=0.5, P_{2, sim}=0.5, P_{3, sim}=-0.5, P_{4, sim}=-0.5, P_{5, sim}=0.5$として, 変更したものをシミュレータとして用いる.
  なお, 初期の$c_{1}, c_{2}$はともに30 $^\circ$Cとする.
  1秒ごとに$f \gets f+\textrm{Rand}(-50.0, 50.0)$ ($\textrm{Rand}(a, b)$は$[a, b]$のランダムな値である, また, $f$は最小値10 N, 最大値200 Nとして制限をかけている)で$f$を更新しながら, その$f$をこのシミュレータに対して送る.
  $c_{2}, f$のみを観測データとしてシミュレータから得ることができる.
  これに対して現在, デフォルトの$P_{\{1, 2, 3, 4, 5\}}=0$のモデルを持っているとする.
  このモデルを使って, 観測された$f$と$c_{2}$から$c_{1}$を推定しておき(同様に$c_{1}$の初期値は30 $^\circ$Cとする), \secref{subsec:online-learning}に示したように, $(c_{1}, c_{2}, f)$からオンライン学習を実行する.
  これにより, 現在のモデルをシミュレータに近づけていく.
  シミュレータと現在のモデルのそれぞれのパラメータの差の絶対値$|P_{\{1, 2, 3, 4, 5\}, sim}-P_{\{1, 2, 3, 4, 5\}}|$とそれら5つのパラメータのRMSEの遷移を3600秒にわたって\figref{figure:simulation-learning}に示す.
  それぞれのパラメータが徐々にシミュレータの値に近づいていくことがわかり, 約1200秒付近で収束している.
  $P_{5}$の誤差が下がり切っていないが, これは$P_{5}$を更新するのに優位な$c_{2}$の温度変化が得られなかったためであると考える.

  また, 3600秒間オンライン学習を実行した後に, 上記と同様に$f$をシミュレータ・学習前のモデル・学習後のモデルに与え, そのときの$f, c_{1}, c_{2}$の遷移を\figref{figure:simulation-evaluation}に示す.
  $c_{1}, c_{2}$について, 学習後のモデルによる温度推定が学習前に比べてシミュレータに近づいていることがわかる.
  教師として$c_{2}$のみを用いて学習が行われているが, $c_{1}$もシミュレータに近づいており, 本手法が有効であることがわかる.
}%

\begin{figure}[t]
  \centering
  \includegraphics[width=1.0\columnwidth]{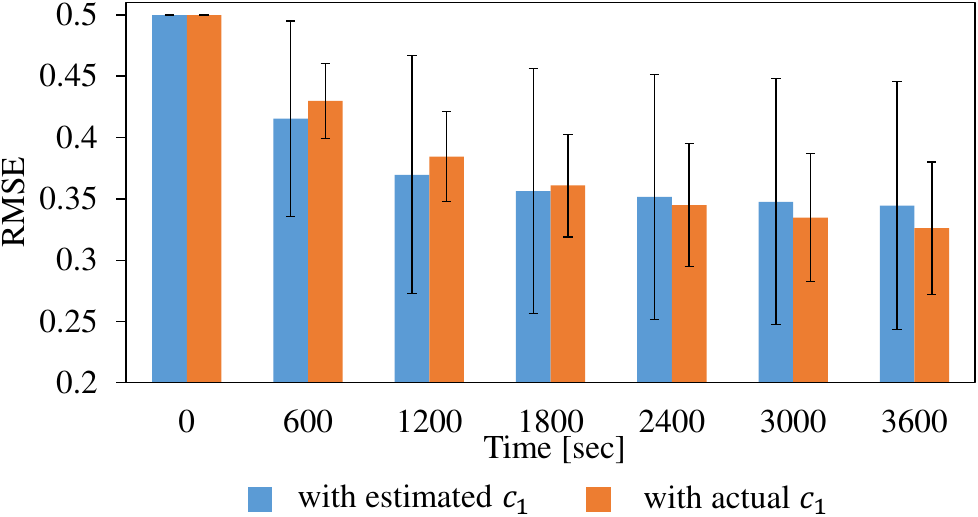}
  \caption{Quantitative evaluation of online learning. The graph shows the transition of RMSE of thermal model parameters between the current model and the simulator with 10 modified parameters during online learning using the estimated or actual $c_{1}$.}
  \label{figure:simulation-quantitative}
  \vspace{-3.0ex}
\end{figure}

\subsubsection{Quantitative Evaluation of Online Learning} \label{subsubsec:simulation-quantitative}
\switchlanguage%
{%
  We constructed 10 parameters of simulators $P_{\{1, 2, 3, 4, 5\}, sim}$ with additional random changes from the default parameters of $P_{\{1, 2, 3, 4, 5\}}$.
  Here, we set the RMSE between default and modified parameters to 0.5.
  We show the average and standard deviation of RMSE of five parameters between the current model and the simulator with 10 modified parameters every 10 minutes of online learning, when executing the same experiment with \secref{subsubsec:simulation-learning}, in \figref{figure:simulation-quantitative}.
  Here, we consider the problem explained in \secref{subsec:online-learning}, that not the actual but only the estimated $c_{1}$ can be obtained.
  Thus, we also show the difference of errors when using the actual and estimated $c_{1}$ as $c^{data}_{1}$ in \secref{subsec:online-learning}, in \figref{figure:simulation-quantitative}.
  As seen from the graph, when using the actual $c_{1}$ as $c^{data}_{1}$, compared with using the estimated one, the error decreased rapidly and its variance was small.
  On the other hand, although there was a certain degree of variance, the current model parameters also gradually became closer to those of the simulator when using the estimated $c_{1}$.
  Also, this result indicates that this study is applicable to various motors with different thermal parameters.
}%
{%
  次に, オンライン学習の定量的な評価を行う.
  まず, デフォルトの$P_{\{1, 2, 3, 4, 5\}}$からランダムに変化を加えた$P_{\{1, 2, 3, 4, 5\}, sim}$を10個作成する.
  このとき, デフォルトの値と変化を加えた値のRMSEが0.5になるようにする.
  \secref{subsubsec:simulation-learning}と同じ実験を10個のパラメータについて実行したときの, 0, 10, 20, 30, 40, 50, 60分後の$|P_{\{1, 2, 3, 4, 5\}, sim}-P_{\{1, 2, 3, 4, 5\}}|$とそれら5つのパラメータのRMSEの平均と分散を\figref{figure:simulation-quantitative}に示す.
  ここで, \secref{subsec:online-learning}で指摘した, $c_{1}$の初期値に実機値ではなく推定値しか得られないという問題についても考える.
  つまり, \secref{subsec:online-learning}における$c^{data}_{1}$を, 実機値としたときと, 推定値としたときの差についても\figref{figure:simulation-quantitative}に示す.
  グラフから, $c^{data}_{1}$は実機値を用いた方が推定値を用いた場合より誤差が速く減っていき, 分散も少ない.
  しかし, 分散はある程度あるものの, 推定値を用いても徐々にモデルのパラメータがシミュレータに近づいていることがわかる.
  よって, 本手法が有効であることが示された.
}%

\begin{figure}[t]
  \centering
  \includegraphics[width=1.0\columnwidth]{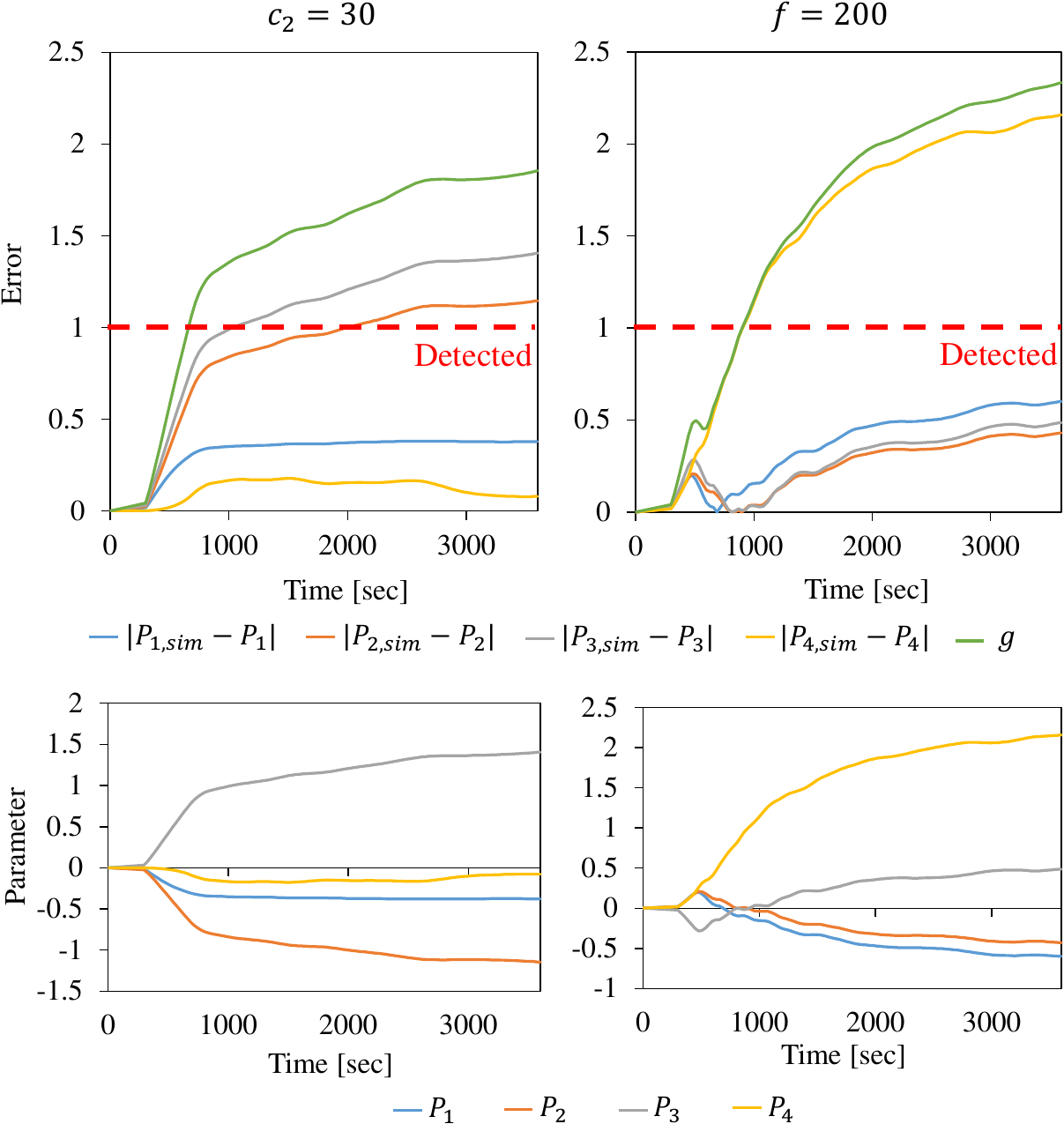}
  \caption{Anomaly detection by learning thermal model parameters. The graph shows the transition of errors in thermal model parameters and $g$ during online learning.}
  \label{figure:simulation-detection}
  \vspace{-3.0ex}
\end{figure}

\subsubsection{Anomaly Detection}
\switchlanguage%
{%
  We executed online learning by setting parameters of the simulator and current model to $P_{\{1, 2, 3, 4, 5\}, sim}=0$.
  Here, we consider two situations.
  One is a situation where the observed $c_{2}$ is always $30$, because of the broken thermal sensor.
  Second is a situation where $f = 200$ is always sent to the simulator although the observed $f$ is correctly updated, because a gear or muscle wire is caught and the muscle does not move as intended.
  We show the transition of parameter errors $|P_{\{1, 2, 3, 4\}, sim}-P_{\{1, 2, 3, 4\}}|$ and $g$ for anomaly detection during online learning for these two situations, in the upper graph of \figref{figure:simulation-detection}.
  Also, in the lower graph of \figref{figure:simulation-detection}, we show the raw change of $P_{\{1, 2, 3, 4\}}$.
  Regarding the former of the two situations, an anomaly was detected after 400 seconds of online learning, and regarding the latter, it was detected after 600 seconds.
  When looking at the values of $P_{\{1, 2, 3, 4\}}$, regarding the former, the change in model parameters to constantly keep $c_{2}=30$ occurred.
  This was due to the increase in the heat escaping from the motor core to the motor housing by making $P_{2}$ small and decreasing heat capacity in motor core, and the decrease in the heat flowing in motor housing from the motor core by making $P_{3}$ large and increasing heat capacity in motor housing.
  Regarding the latter, by increasing $P_{4}$ and inhibiting the escape of heat from the motor housing to the ambient, a situation in which high $c_{2}$ is constantly observed due to $f=200$ was produced.
  Thus, we can detect and interpret anomalies such as a broken thermal sensor, broken gear, and burnout of motors by using the interpretability of thermal model parameters.
}%
{%
  シミュレータのパラメータをデフォルトの$P_{\{1, 2, 3, 4, 5\}, sim}=0$とし, オンライン学習を実行する.
  このとき, 常に観測できる値が$c_{2}=30$で一定になってしまう状態, または, 観測できる$f$は更新される値だが, 実際にはシミュレータに対して$f=200$が常にかかってしまっている状態, の2つを考える.
  前者は温度センサが壊れている状態, 後者はギアや筋が引っ掛かり筋が思ったように動かない状態を想定している.
  このときのオンライン学習によるパラメータ誤差$|P_{\{1, 2, 3, 4\}, sim}-P_{\{1, 2, 3, 4\}}|$の変化, 異常検知の値$g$の遷移を\figref{figure:simulation-detection}の上図に示す.
  また, \figref{figure:simulation-detection}の下図に, $P_{\{1, 2, 3, 4\}}$の変化も同様に示す.
  前者の場合はオンライン学習が開始してから約400秒, 後者は600秒で異常が検知されていることがわかる.
  実際の$P_{\{1, 2, 3, 4\}}$の値を見ると, 前者では$P_{2}$を小さく(モータコアの熱容量を下げる)してモータコアからモータハウジングに逃げる熱の量を増やし, $P_{3}$を上げる(モータハウジングの熱容量を上げる)ことでモータコアからの熱の流入を減らしており, モータハウジング温度を30度で一定に保つための変化が起きていることがわかる.
  また後者では, $P_{4}$を増やしてモータハウジングの熱が外部に流れるのを防ぎ, $f=200$で一定のため常に高い$c_{2}$が観測され続ける状態を再現している.
  このように, 温度センサが壊れた, ギアが壊れてモータが回らない, モータが焼けて動かずに温度のみが上昇していく, 等の異常を検知でき, かつその状況をパラメータの解釈性から読み取ることができる可能性がある.
}%

\begin{figure}[t]
  \centering
  \includegraphics[width=1.0\columnwidth]{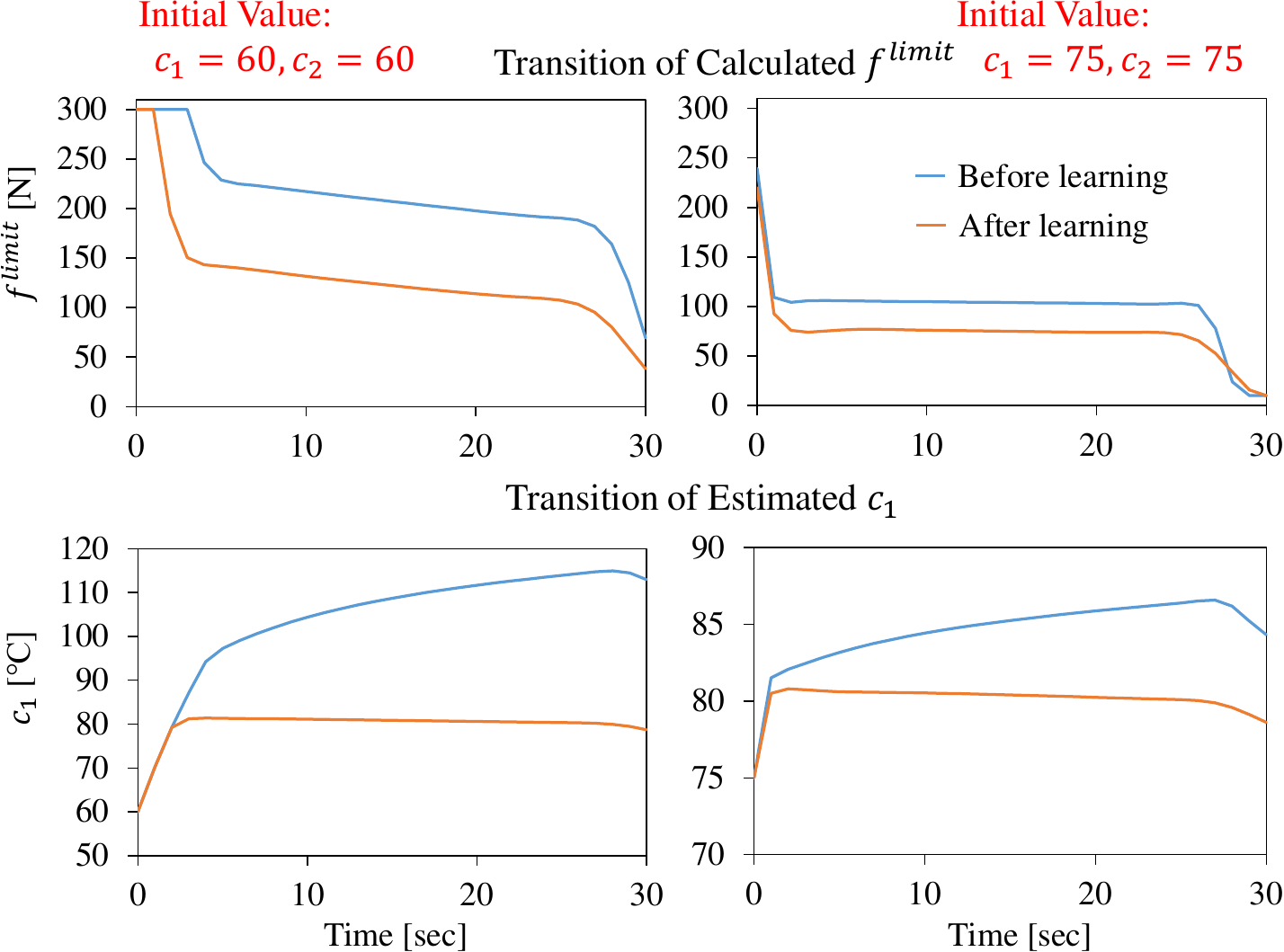}
  \caption{Calculation of the optimized muscle tension to control $c_{1}$ and its evaluation. The graph shows the transition of the calculated $f^{limit}$ and observed $c_{1}$ with initial $c_{1}$ and $c_{2}$ of 60 or 75.}
  \label{figure:simulation-control}
  \vspace{-3.0ex}
\end{figure}

\subsubsection{Thermal Controller}
\switchlanguage%
{%
  Regarding the model of \secref{subsubsec:simulation-learning} before and after online learning, when both the current $c_{1}$ and $c_{2}$ are 60 or 75, we observed the calculated $f^{limit}_{[k, k+N_{control}-1]}$.
  Also, we compared the transition of $c_{1}$ when sending the calculated $f^{limit}_{[k, k+N_{control}-1]}$ to the simulator, the model before online learning, and the model after online learning.
  As shown in \figref{figure:simulation-control}, regarding both the initial temperatures of 60 and 75, the calculated $f^{limit}$ is high at first, gradually decreases, and is finally kept constant.
  Also, while $c_{1}$ achieved $c^{max}_{1}$ accurately when using the model after online learning, $c_{1}$ largely exceeded $c^{max}_{1}$ when using the model before online learning.
  Thus, online learning of the thermal model is effective for the thermal controller, and this method can always keep $c_{1}$ below $c^{max}_{1}$.
}%
{%
  \secref{subsubsec:simulation-learning}の学習前のモデル・学習後のモデルについて, $c_{1}, c_{2}$が現在両者とも60, または75のときに, どのような$f^{limit}_{[k, k+N_{control}-1]}$が計算されるのかを観察する.
  また, 実際に計算された$f^{limit}_{[k, k+N_{control}-1]}$をシミュレータに入力したときの$c_{1}$の遷移を学習前と学習後のモデルで比較する.
  \figref{figure:simulation-control}に示すように, 初期温度が60, 75の両者で同様に, 最初は高く, 徐々に値が落ちて最終的になだらかで一定となるような$f^{limit}$が計算されていることがわかる.
  また, $c_{1}$の遷移は, 学習後のモデルでは正しく最大値に設定した$c^{max}_{1}$を実現できているのに対して, 学習前のモデルでは$c_{1}$が$c^{max}_{1}$を大きく超えてしまっていることがわかる.
  よって, 温度制限制御を行うにあたって温度モデルのオンライン学習は有効であり, また, 本手法によってモータコアの温度を$c^{max}_{1}$以下に常に保ち続けることができることがわかった.
}%

\begin{figure}[t]
  \centering
  \includegraphics[width=1.0\columnwidth]{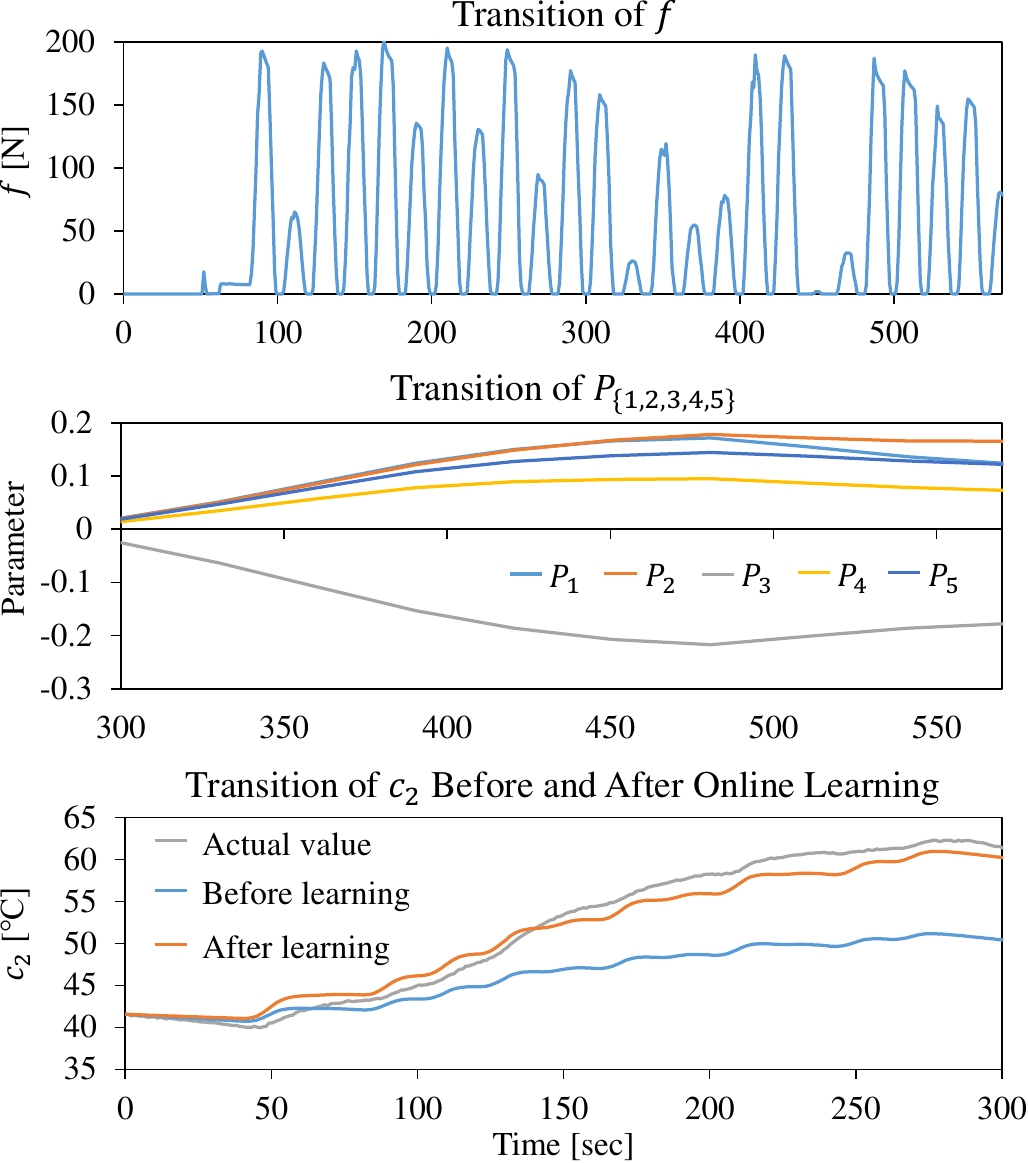}
  \caption{Online learning experiment with one actual muscle actuator. The upper graph shows the transition of the applied $f$ during online learning, the middle graph shows the transition of $P_{\{1, 2, 3, 4, 5\}}$, and the lower graph shows the transition of the estimated $c_{2}$ before and after online learning.}
  \label{figure:onemuscle-learning}
  \vspace{-3.0ex}
\end{figure}

\begin{figure}[t]
  \centering
  \includegraphics[width=1.0\columnwidth]{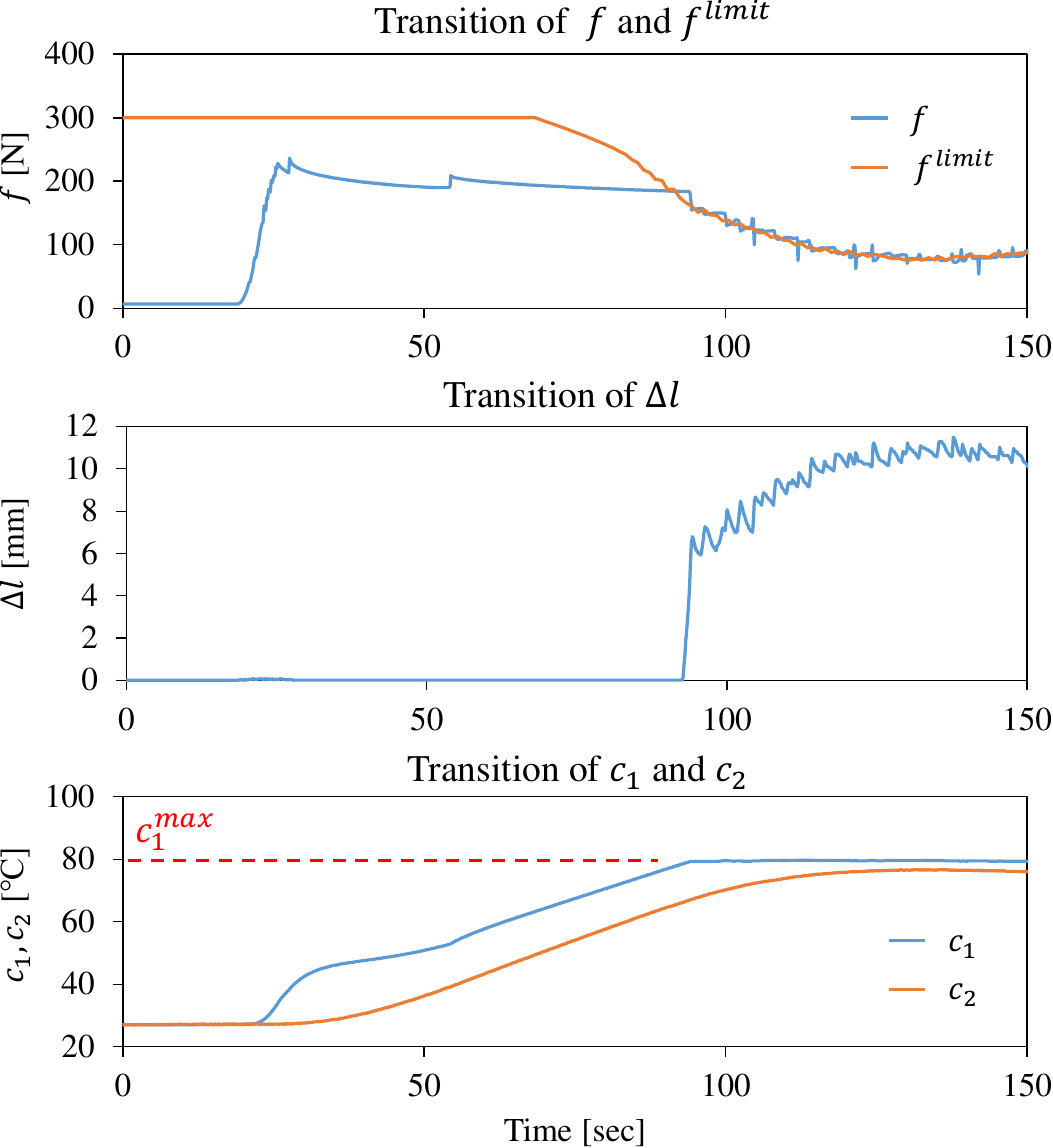}
  \caption{Thermal control experiment with one actual muscle actuator. The upper graph shows the transition of $f$ and calculated $f^{limit}$, the middle graph shows the transition of $\Delta{l}$, and the lower graph shows the transition of $c_{1}$ and $c_{2}$.}
  \label{figure:onemuscle-control}
  \vspace{-3.0ex}
\end{figure}

\subsection{Actual Experiment with One Muscle Actuator}
\subsubsection{Online Learning}
\switchlanguage%
{%
  We fixed the end position of the muscle, pulled $\bm{l}^{ref}$ by $\textrm{Rand}(-16, 0)$ [mm] over 10 seconds, elongated it over 10 seconds, and repeated these procedures.
  We show the transition of $f$ and $P_{\{1, 2, 3, 4, 5\}}$ during online learning in \figref{figure:onemuscle-learning}.
  After the online learning, when changing $\bm{l}^{ref}$ as stated above again, we compared the obtained $c_{2}$ from the actual muscle actuator and the estimated $c_{2}$ using the model before and after online learning, in the lower graph of \figref{figure:onemuscle-learning}.
  After online learning, the estimated value became closer to the actual sensor value than before online learning.
  Although the comparison of $c_{1}$ is difficult because the actual $c_{1}$ cannot be obtained, from \figref{figure:simulation-evaluation} in \secref{subsubsec:simulation-learning}, the estimated $c_{1}$ is considered to become accurate by making $c_{2}$ correct.
  Therefore, this method is also applicable to the actual muscle actuator.
}%
{%
  一本の実際の筋アクチュエータを使って, オンライン学習を評価する.
  筋の末端の位置を固定したうえで, 10秒かけて$\textrm{Rand}(-16, 0)$ [mm]の筋長を送って筋を緊張させ, 10秒かけて戻すことを繰り返す.
  その際の筋張力変化, 同時にオンライン学習を実行した際の温度モデルパラメータの推移を\figref{figure:onemuscle-learning}に示す.
  その後, もう一度同様に筋長を変化させたときに, 実機から得られた$c_{2}$, 学習前と学習後におけるモデルを使って推定した$c_{2}$を\figref{figure:onemuscle-learning}に比較する.
  学習後は, 学習前に比べて大きく推定値が実機に近づいていることがわかる.
  $c_{1}$の実機値は得ることができないため比較は難しいが, \secref{subsubsec:simulation-learning}の\figref{figure:simulation-evaluation}から$c_{2}$の推定が近づくことで$c_{1}$の推定もより正しくなっていると考えられる.
  そのため, 実機においても本手法が適用可能なことがわかった.
}%

\subsubsection{Thermal Controller}
\switchlanguage%
{%
  We verified whether $c_{1}$ can be always kept below $c^{max}_{1}$ by using the methods of \secref{subsec:thermal-controller} and \secref{subsec:tension-limiter}, and the updated model.
  We fixed the muscle endpoint, sent -16 mm for $\bm{l}^{ref}$, and generated $f$ of about 200 N.
  At the same time, by the muscle tension limiter of \secref{subsec:tension-limiter}, $f$ was restricted by $f^{limit}$.
  We show the transition of the observed $f$, calculated $f^{limit}$, $\Delta{l}$, estimated $c_{1}$, and observed $c_{2}$ in \figref{figure:onemuscle-control}.
  $c_{1}$ gradually increased by keeping high $f$, and $f^{limit}$ began to decrease below $f^{max}$ in about 70 seconds.
  In about 90 seconds, $f$ exceeded $f^{limit}$, and $f$ was inhibited by gradually increasing $\Delta{l}$.
  We can see that $c_{1}$ constantly achieves $c^{max}_{1}$ by this control.
  Although $f^{max}$ should be about 100 N in order to keep $c_{1}$ below $c^{max}_{1}$, by using this study, large $f$ is ordinarily outputted but the output can be restricted by watching $c_{1}$.
}%
{%
  学習されたモデルを用いて, \secref{subsec:thermal-controller}, \secref{subsec:tension-limiter}を用いて, $c_{1}$を常に$c^{max}_{1}$以下に保つことができるかを検証する.
  筋長として-16 mmを送り, 約200 Nの筋張力を発生させる.
  これと同時に筋張力制限制御によって, 筋張力を制限させる.
  その際の$f$, 計算された$f^{limit}$, 筋弛緩量$\Delta{l}$, 推定された$c_{1}$, 実測値$c_{2}$の遷移を\figref{figure:onemuscle-control}に示す.
  高い筋張力をかけ続けることで$c_{1}$は徐々に上がっていき, 約70 secの時点で計算された$f^{limit}$の値が$f^{max}$よりも下がり始める.
  そして, 約90 secのところで$f$が$f^{limit}$を越え, $\Delta{l}$が徐々に増え, 筋張力が抑制されている.
  この制御により, $c_{1}$がぴったりと$c^{max}_{1}$を実現し続けていることがわかる.
  モータコア温度を$c^{max}_{1}$以下に抑えるためには$f^{max}$を100 N程度に抑える必要があるが, 本研究により, 通常はより大きな力を出し, モータコア温度を見ながら出力を下げていくことができることがわかった.
}%

\subsection{Application to the Musculoskeletal Humanoid}
\begin{figure}[t]
  \centering
  \includegraphics[width=1.0\columnwidth]{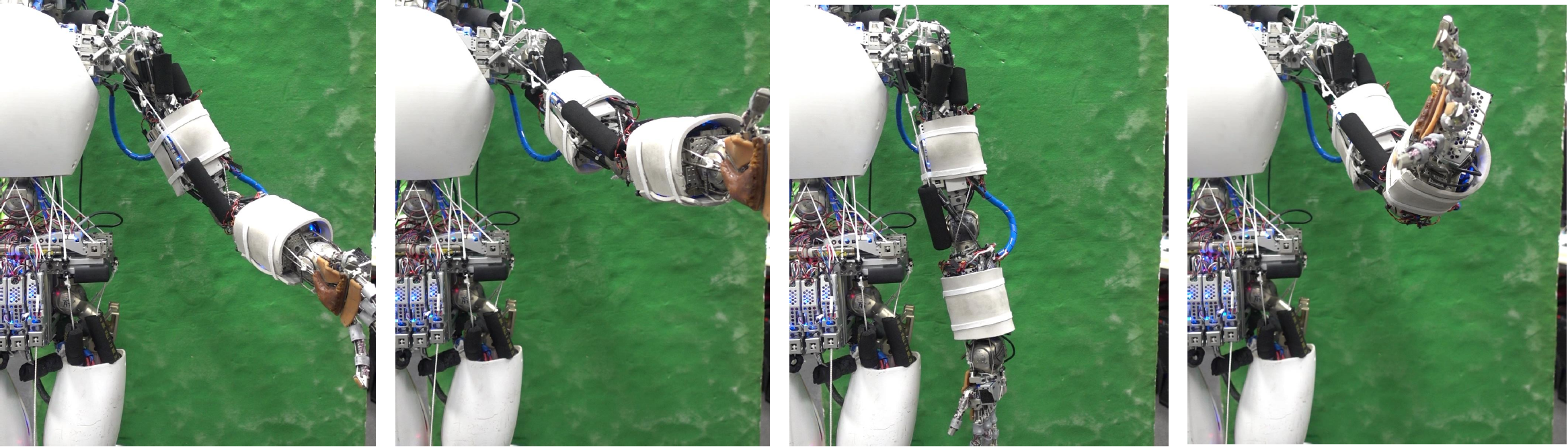}
  \vspace{-3.0ex}
  \caption{Random movements of the left arm of the musculoskeletal humanoid Musashi \cite{kawaharazuka2019musashi}.}
  \label{figure:musculoskeletal-appearance}
  \vspace{-1.0ex}
\end{figure}

\begin{figure}[t]
  \centering
  \includegraphics[width=1.0\columnwidth]{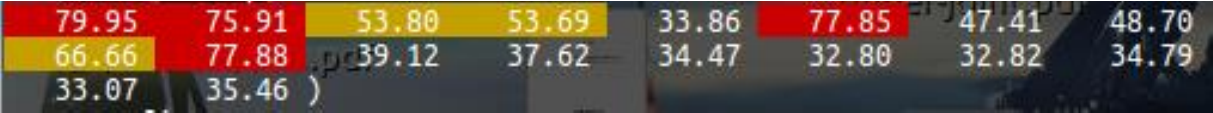}
  \vspace{-3.0ex}
  \caption{Monitoring of $c_{1}$ with colors for manual thermal management. Each value shows $c_{1}$ included in the left arm of Musashi. When the value is colored red, the temperature is high ($>$70 $^\circ$C), and when the value is colored yellow, the temperature is slightly high ($>$50 $^\circ$C).
  }
  \label{figure:musculoskeletal-display}
  \vspace{-3.0ex}
\end{figure}

\begin{figure}[t]
  \centering
  \includegraphics[width=1.0\columnwidth]{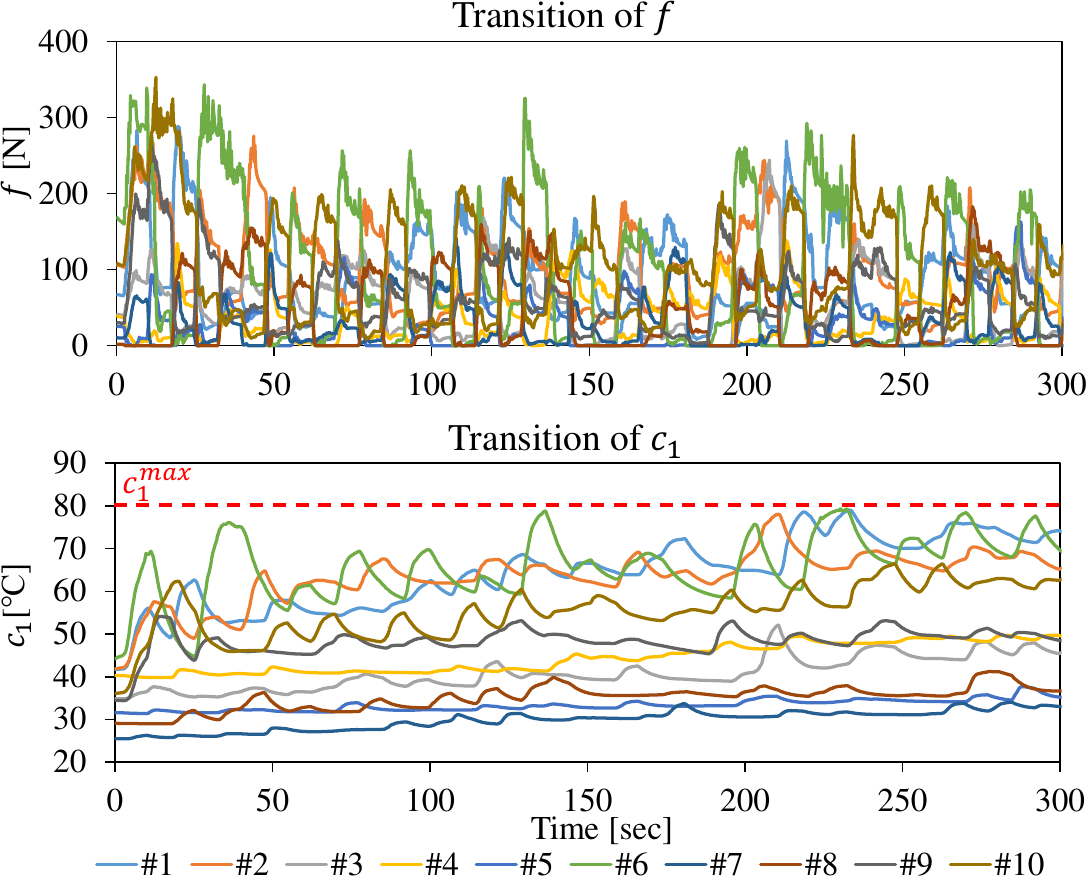}
  \vspace{-3.0ex}
  \caption{Transition of $f$ and $c_{1}$ during random movements of the left arm of Musashi.}
  \label{figure:musculoskeletal-control}
  \vspace{-3.0ex}
\end{figure}

\switchlanguage%
{%
  We consider how this study contributes to movements of the actual robot, by applying to the muscles of the left arm of Musashi \cite{kawaharazuka2019musashi}.
  While executing online learning, thermal estimation, thermal controller, and muscle tension limiter, we sent random joint angles to the robot as shown in \figref{figure:musculoskeletal-appearance}.
  We moved the robot so that high $f$ is constantly generated by using variable stiffness control \cite{kawaharazuka2019longtime}.
  The important point here is not only limiting $c_{1}$ but also displaying the estimated $c_{1}$ with colors, as shown in \figref{figure:musculoskeletal-display}.
  By monitoring $c_{1}$ updated accurately by online learning, we can rapidly find anomaly during experiments.
  Also, we show the transition of $f$ and $c_{1}$ of 10 muscles \#1-\#10 contributing to the shoulder and elbow of Musashi, in \figref{figure:musculoskeletal-control}.
  The main muscles with high tension were \#1, \#2, \#6, and \#8.
  Although high $f$ of about 350 N occurred at the initial stage, as $c_{1}$ increased, the maximum $f$ was inhibited to about 250 N.
  Especially, the $c_{1}$ of \#1, \#2, and \#6 vibrated around below 80, and we can see that the continuous movements of Musashi are achieved by the developed thermal controller.
  Therefore, this study is applicable to multiple muscle actuators, and is effective in movements of the actual robot.

  In applying this study to the 36 muscles of both arms of Musashi, the calculation of the thermal estimator, online learning, and thermal controller takes 5, 200, and 200 msec, respectively.
  By rewriting the current program written in Python using C++ or modifying parameters of $N_{seq}$, $N_{control}$, and $N_{batch}$, this study can work at a higher frequency.
}%
{%
  筋骨格ヒューマノイドMusashi \cite{kawaharazuka2019musashi}の左腕の筋に対して本研究を適用し, 実際のロボットの動きの中でどのように本研究が寄与するかを考察する.
  オンライン学習・モータコア温度推定・筋張力制限制御を走らせながら, ランダムな関節角度を送り続けることを行う.
  この際に, \cite{kawaharazuka2019longtime}の可変剛性制御を用い, 高い剛性を実現する一方で大きな力が常にかかり続けるような状態で実験を行う.
  ここで重要なのは, $c_{1}$を制限するだけでなく, 常に推定されたモータコア温度を\figref{figure:musculoskeletal-display}に示すように, 色をつけて常に監視することである.
  オンライン学習によってより正確になったモータコア温度を監視し, 実験中にいち早く異常に気づけるようにすることが重要である.
  また, この際の$f, c_{1}$の推移を, 肩・肘に寄与する10本の筋\#1-\#10に関して, \figref{figure:musculoskeletal-control}に示す.
  主に高い筋張力を発しているのは\#1, \#2, \#6, \#8である.
  動作初期は最大で350 N程度の強い力がかかることがあるが, $c_{1}$が上がるにつれ, 最大筋張力が250 N程度まで筋張力が抑えられるように変化している.
  特に, \#1, \#2, \#6の筋は$c_{1}$が80度付近までいっては戻るような動きをしており, 筋張力制限制御により, 継続的な動作が可能となっていることがわかる.
  このように, 本研究を複数のアクチュエータに対して適用し, 実際の動作の中でも有効であることが示された.

  最後に, 例えばMusashiの両腕36自由度に本研究を適用した場合, 温度推定は5 msec, 温度制御は200 msec, オンライン学習は200 msec程度の時間が必要になる.
  本研究のプログラムは全てPythonで書かれているため, コンパイル言語で書き直す, または$N_{seq}$や$N_{control}$, $N_{batch}$などを変更することで, より速いHzで動作させることが可能である.
}%

\section{CONCLUSION} \label{sec:conclusion}
\switchlanguage%
{%
  In this study, for smart management of motor temperature, we proposed an online learning method of thermal model parameters, estimation of motor core temperature, anomaly detection of motors, and control of motor core temperature by optimization.
  By referring to the observed motor housing temperature, the thermal model parameters can be updated online.
  Anomaly detection is easily enabled by using not a neural network but explicit parameters of the thermal model.
  The time series maximum output to achieve the maximum motor core temperature can be calculated by backpropagation, and motor core temperature can be managed by using it.
  Finally, we verified the effectiveness of this study in a simulation and in the actual robot, and achieved its application to the musculoskeletal humanoid.

  In the current form, the thermal management can change the whole-body motion of the robot, and this can cause severe trouble for walking, task execution, etc.
  In future works, we would like to embed the calculated maximum output value in motion planning.
}%
{%
  本研究では, 賢いモータ温度管理のための, 温度モデルパラメータのオンライン学習, そのモデルを使ったモータコア温度推定, 異常検知, モータコア温度管理のための出力制限制御について提案した.
  モータハウジング温度のマッチングによって温度モデルパラメータをオンラインで学習させることができる.
  また, ニューラルネットワークの重みではなく明示的に解釈可能なパラメータを用いることで, 異常検知が可能となる.
  モータコア温度を最大値に保つための最大出力の時系列も誤差逆伝播法を用いて得ることが可能であり, これによってモータコア温度を管理することが可能となる.
  最後に, 本研究の有効性をシミュレーション・実機により示し, 筋骨格ヒューマノイドへの適用も実現した.

  今後は, 得られた出力制限値を動作計画に組み込んでいきたい.
}%

{
  \bibliographystyle{IEEEtran}
  \bibliography{main}

\begin{thebibliography}{10}
\providecommand{\url}[1]{#1}
\csname url@rmstyle\endcsname
\providecommand{\newblock}{\relax}
\providecommand{\bibinfo}[2]{#2}
\providecommand\BIBentrySTDinterwordspacing{\spaceskip=0pt\relax}
\providecommand\BIBentryALTinterwordstretchfactor{4}
\providecommand\BIBentryALTinterwordspacing{\spaceskip=\fontdimen2\font plus
\BIBentryALTinterwordstretchfactor\fontdimen3\font minus
  \fontdimen4\font\relax}
\providecommand\BIBforeignlanguage[2]{{%
\expandafter\ifx\csname l@#1\endcsname\relax
\typeout{** WARNING: IEEEtran.bst: No hyphenation pattern has been}%
\typeout{** loaded for the language `#1'. Using the pattern for}%
\typeout{** the default language instead.}%
\else
\language=\csname l@#1\endcsname
\fi
#2}}

\bibitem{urata2010design}
J.~Urata, Y.~Nakanishi, K.~Okada, and M.~Inaba, ``{Design of high torque and
  high speed leg module for high power humanoid},'' in \emph{Proceedings of the
  2010 IEEE/RSJ International Conference on Intelligent Robots and Systems},
  2010, pp. 4497--4502.

\bibitem{sevinchan2018thermal}
E.~Sevinchan, I.~Dincer, and H.~Lang, ``{A review on thermal management methods
  for robots},'' \emph{Applied Thermal Engineering}, vol. 140, pp. 799--813,
  2018.

\bibitem{erez2013mpc}
T.~Erez, K.~Lowrey, Y.~Tassa, V.~Kumar, S.~Kolev, and E.~Todorov, ``{An
  integrated system for real-time model predictive control of humanoid
  robots},'' in \emph{Proceedings of the 2013 IEEE-RAS International Conference
  on Humanoid Robots}, 2013, pp. 292--299.

\bibitem{diehl2006shooting}
M.~Diehl, H.~G. Bock, H.~Diedam, and P.~B. Wieber, \emph{{Fast Direct Multiple
  Shooting Algorithms for Optimal Robot Control}}.\hskip 1em plus 0.5em minus
  0.4em\relax Springer Berlin Heidelberg, 2006, pp. 65--93.

\bibitem{noda2014thermal}
S.~Noda, M.~Murooka, S.~Nozawa, Y.~Kakiuchi, K.~Okada, and M.~Inaba, ``{Online
  maintaining behavior of high-load and unstable postures based on whole-body
  load balancing strategy with thermal prediction},'' in \emph{Proceedings of
  the 2014 IEEE International Conference on Automation Science and
  Engineering}, 2014, pp. 1166--1171.

\bibitem{urata2008thermal}
J.~Urata, T.~Hirose, Y.~Namiki, Y.~Nakanishi, I.~Mizuuchi, and M.~Inaba,
  ``{Thermal control of electrical motors for high-power humanoid robots},'' in
  \emph{Proceedings of the 2008 IEEE/RSJ International Conference on
  Intelligent Robots and Systems}, 2008, pp. 2047--2052.

\bibitem{kumagai2014highload}
I.~Kumagai, S.~Noda, S.~Nozawa, Y.~Kakiuchi, K.~Okada, and M.~Inaba, ``{Whole
  body joint load reduction control for high-load tasks of humanoid robot
  through adapting joint torque limitation based on online joint temperature
  estimation},'' in \emph{Proceedings of the 2014 IEEE-RAS International
  Conference on Humanoid Robots}, 2014, pp. 463--468.

\bibitem{gravato2010ecce1}
H.~G. Marques, M.~J{\"a}ntsh, S.~Wittmeier, O.~Holland, C.~Alessandro,
  A.~Diamond, M.~Lungarella, and R.~Knight, ``{ECCE1: the first of a series of
  anthropomimetic musculoskeletal upper torsos},'' in \emph{Proceedings of the
  2010 IEEE-RAS International Conference on Humanoid Robots}, 2010, pp.
  391--396.

\bibitem{kawaharazuka2019musashi}
K.~Kawaharazuka, S.~Makino, K.~Tsuzuki, M.~Onitsuka, Y.~Nagamatsu, K.~Shinjo,
  T.~Makabe, Y.~Asano, K.~Okada, K.~Kawasaki, and M.~Inaba, ``{Component
  Modularized Design of Musculoskeletal Humanoid Platform Musashi to
  Investigate Learning Control Systems},'' in \emph{Proceedings of the 2019
  IEEE/RSJ International Conference on Intelligent Robots and Systems}, 2019,
  pp. 7294--7301.

\bibitem{kozuki2016sweat}
T.~Kozuki, H.~Toshinori, T.~Shirai, S.~Nakashima, Y.~Asano, Y.~Kakiuchi,
  K.~Okada, and M.~Inaba, ``{Skeletal structure with artificial perspiration
  for cooling by latent heat for musculoskeletal humanoid Kengoro},'' in
  \emph{Proceedings of the 2016 IEEE/RSJ International Conference on
  Intelligent Robots and Systems}, 2016, pp. 2135--2140.

\bibitem{rumelhart1986bptt}
D.~E. Rumelhart, G.~E. Hinton, and R.~J. Williams, ``{Learning Internal
  Representations by Error Propagation},'' in \emph{Parallel Distributed
  Processing: Explorations in the Microstructure of Cognition, Vol. 1}, D.~E.
  Rumelhart, J.~L. McClelland, and C.~PDP Research~Group, Eds.\hskip 1em plus
  0.5em minus 0.4em\relax Cambridge, MA, USA: MIT Press, 1986, pp. 318--362.

\bibitem{pascanu2013clipping}
R.~Pascanu, T.~Mikolov, and Y.~Bengio, ``{On the Difficulty of Training
  Recurrent Neural Networks},'' in \emph{Proceedings of the 30th International
  Conference on Machine Learning}, 2013, pp. 1310--1318.

\bibitem{jantsch2012computed}
M.~J{\"a}ntsch, S.~Wittmeier, K.~Dalamagkidis, and A.~Knoll, ``{Computed muscle
  control for an anthropomimetic elbow joint},'' in \emph{Proceedings of the
  2012 IEEE/RSJ International Conference on Intelligent Robots and Systems},
  2012, pp. 2192--2197.

\bibitem{asano2015sensordriver}
Y.~Asano, T.~Kozuki, S.~Ookubo, K.~Kawasaki, T.~Shirai, K.~Kimura, K.~Okada,
  and M.~Inaba, ``{A Sensor-driver Integrated Muscle Module with High-tension
  Measurability and Flexibility for Tendon-driven Robots},'' in
  \emph{Proceedings of the 2015 IEEE/RSJ International Conference on
  Intelligent Robots and Systems}, 2015, pp. 5960--5965.

\bibitem{kawaharazuka2017forearm}
K.~Kawaharazuka, S.~Makino, M.~Kawamura, Y.~Asano, Y.~Kakiuchi, K.~Okada, and
  M.~Inaba, ``{Human Mimetic Forearm Design with Radioulnar Joint using
  Miniature Bone-muscle Modules and its Applications},'' in \emph{Proceedings
  of the 2017 IEEE/RSJ International Conference on Intelligent Robots and
  Systems}, 2017, pp. 4956--4962.

\bibitem{kawaharazuka2019longtime}
K.~Kawaharazuka, K.~Tsuzuki, S.~Makino, M.~Onitsuka, Y.~Asano, K.~Okada,
  K.~Kawasaki, and M.~Inaba, ``{Long-time Self-body Image Acquisition and its
  Application to the Control of Musculoskeletal Structures},'' \emph{IEEE
  Robotics and Automation Letters}, vol.~4, no.~3, pp. 2965--2972, 2019.

\end{thebibliography}
}

\end{document}